\documentclass{article}

\usepackage[numbers,sort&compress]{natbib}
\usepackage[final,nonatbib]{neurips_2023}

\usepackage[utf8]{inputenc} %
\usepackage[T1]{fontenc}    %
\usepackage{hyperref}       %
\usepackage{url}            %
\usepackage{booktabs}       %
\usepackage{amsfonts}       %
\usepackage{nicefrac}       %
\usepackage{microtype}      %
\usepackage[dvipsnames]{xcolor}         %
\usepackage{graphicx}
\usepackage{newtxmath}
\usepackage{wrapfig}
\usepackage{adjustbox}
\usepackage{multirow}
\usepackage{multicol}
\usepackage{colortbl}
\usepackage[symbol]{footmisc}
\usepackage{listings}

\usepackage[capitalize]{cleveref}
\crefname{section}{Sec.}{Secs.}
\Crefname{section}{Section}{Sections}
\Crefname{table}{Table}{Tables}
\crefname{table}{Tab.}{Tabs.}

\title{Exploring Question Decomposition for Zero-Shot VQA}

\author{%
Zaid Khan$^{1}$ \quad Vijay Kumar BG$^{2}$ \quad Samuel Schulter$^2$ \quad Manmohan Chandraker$^{2,3}$ \quad Yun Fu$^2$ \\
$^1$Northeastern University \quad $^2$NEC Laboratories America \quad $^3$UC San Diego\\
\texttt{\{khan.za,yun.fu\}@northeastern.edu}\\
\texttt{\{vijay.kumar, samuel, manu\}@nec-labs.com}\\
}

\begin{document}

\maketitle

\begin{abstract}
Visual question answering (VQA) has traditionally been treated as a single-step task where each question receives the same amount of effort, unlike natural human question-answering strategies. We explore a question decomposition strategy for VQA to overcome this limitation. We probe the ability of recently developed large vision-language models to use human-written decompositions and produce their own decompositions of visual questions, finding they are capable of learning both tasks from demonstrations alone.
However, we show that naive application of model-written decompositions can hurt performance.
We introduce a model-driven \textit{selective decomposition} approach for second-guessing predictions and correcting errors, and validate its effectiveness on eight VQA tasks across three domains, showing consistent improvements in accuracy, including improvements of $>20$\% on medical VQA datasets and boosting the zero-shot performance of BLIP-2 above chance on a VQA reformulation of the challenging Winoground task. 
Project Site: \url{https://zaidkhan.me/decomposition-0shot-vqa/}

\end{abstract}

\section{Introduction}
On a question-answering test, humans are able to answer some questions in a single step, while other questions require potential deliberation and second-guessing.
Visual question answering (VQA) \cite{Goyal2016MakingTV,Khan2023QHT,Agrawal2015VQAVQ} has traditionally been treated as a single-step task.
Models only get one chance for each question, and each question receives equal amounts of computation.
This is incongruent to the natural human approach to such tasks, where simple perceptual questions are quickly answered, while harder reasoning questions are allocated more time and computation.

The emergence of task decomposition techniques for large language models (LLMs) \cite{Mialon2023AugmentedLM} is a potential solution to this incongruency.
Task decomposition techniques \textit{prompt} a LLM to break down an initial complex task into simpler subtasks that can each be solved independently.
However, VQA has not benefited from advances in task decomposition techniques for two reasons.
First, many task decomposition techniques \cite{Wei2022ChainOT,Wang2022SelfConsistencyIC} have only been effective in the regime of very large unimodal LLMs with parameters in the 30B+ range, while the LLMs underlying vision-language models are typically much smaller, only recently reaching $\approx13$b parameters for publicly available models\cite{blip2,Liu2023VisualIT,Zhu2023MiniGPT4EV}.
Second, existing methods for prompting vision-language models (VLMs) during VQA tasks focus on other use cases, such as providing more examples of the input task \cite{Yang2021AnES} or more information about the image \cite{Guo2022FromIT}.
Given the recent emergence of multi-billion scale VLMs, our main research question is: 
\begin{quote}
    \textit{Can multi-billion scale vision-language models benefit by approaching reasoning-heavy VQA as a two-step rather than a single-step problem using decomposition?}
\end{quote}
To this end, we explore a form of task decomposition called \textit{question decomposition} as a strategy for zero-shot visual question answering with large VLMs.
Although question decomposition has been explored for specific unimodal QA\cite{Perez2020UnsupervisedQD,Khot2021TextMN,Patel2022IsAQ}, it has not been explored as a strategy for multimodal tasks such as VQA with emerging large VLMs \cite{blip2,Liu2023VisualIT,Zhu2023MiniGPT4EV,magma,koh2023grounding}, and little is known about the in-context learning ability of emerging large VLMs.

First, we probe the in-context learning ability \cite{chan_data_2022,Min2022RethinkingTR,Xie2021AnEO} of both LMs and VLMs to exploit oracular question decompositions written by humans.
We design experiments to understand whether models can learn to use decompositions without explicit training, and whether they are merely exploiting keywords and surface statistics when they use decompositions.
Second, we conduct a series of experiments, again using in-context learning, to understand how well models can \textit{produce} decompositions that correct the errors of a fixed VQA model.
Last, we propose and study an entirely model-driven closed-loop approach mimicking a simplified form of a classic human second-guessing strategy: second-guess answers based on how confident you are about them.
We conduct experiments across three domains (art, natural images, medical), eight datasets, three model families, and model sizes ranging from 80M to 11B parameters. Our contributions can be listed as follows: 
\begin{enumerate}
    \item We experimentally demonstrate that large VLMs based on instruction-tuned LLMs can use decompositions to improve their predictions without any training, and are not merely exploiting changes in word statistics introduced by the decomposition. (\cref{sec:oraclular-exprs})
    \item We quantitatively show that generative, instruction-tuned language models are capable of writing effective decompositions zero-shot, without task-specific training. (\cref{sec:decomp-production})
    \item We find that applying decomposition naively to every question instance harms performance rather than helps (\cref{fig:decomposition_curve}), and propose \textit{selective decomposition} (\cref{code:selective-decomposition}), a modular, model-agnostic, training-free strategy that treats VQA as a two-step task. (\cref{sec:selective_decomposition})
    \item We apply selective decomposition to a testbed of 8 datasets and show that it consistently improves performance (\cref{tab:natural_domain_net_gain,tab:other_domain_net_gain}), with gains of $>20\%$ on medical VQA datasets\cite{slake,vqa_rad,pathvqa}, and boosts the performance of BLIP-2\cite{blip2} above chance on the Winoground\cite{winoground} benchmark when formulated as a VQA task. (\cref{sec:selective_decomposition}).
\end{enumerate}

\begin{figure}
    \centering
    \includegraphics[width=\linewidth]{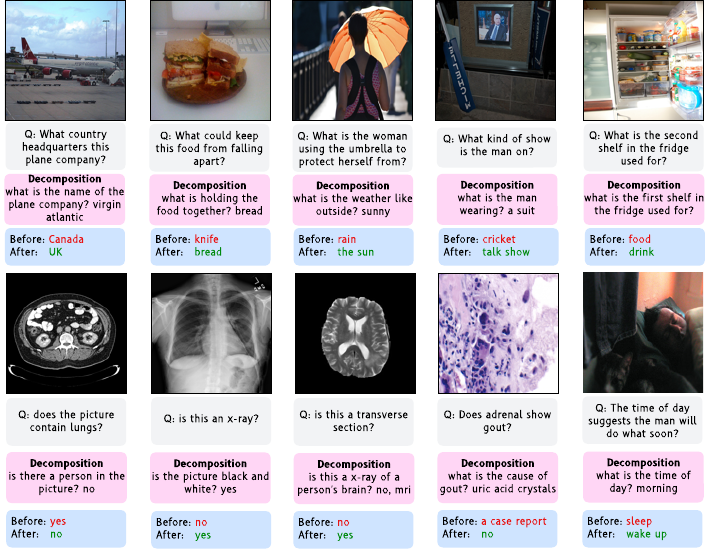}
    \caption{\small{Model-produced decompositions and their error correcting effects. The decompositions and before/after answers shown above were produced by prompting BLIP-2 models based on FLAN-T5 to produce a subquestion, answering the subquestion with the model and feeding the question and answered subquestion back to the model: it is correcting itself. \textcolor{red}{Before} answers are wrong and \textcolor{ForestGreen}{After} answers are correct.}}
    \label{fig:qualitative_examples}
    \vspace{-4mm}
\end{figure}

\section{Background}
\subsection{Problem Setting}
\label{sec:problem-setting}
In zero-shot VQA, a model $f:v,q \rightarrow a$ is given an image $v$, a question $q$, and outputs an answer, $a$. 
Unlike traditional VQA, the model $f(\cdot)$ has never seen $v,q,a$ triplets.
In practice, such a setting often occurs when $f(\cdot)$ is a foundation model that contains several billion parameters and has undergone large scale pretraining.
It is undesirable to retrain such an $f(\cdot)$  on visual question answering pairs specifically, both for reasons of computational convenience and because finetuning can degrade robustness\cite{Wortsman2021RobustFO}. 
The most common case is that $f(\cdot)$ is an autoregressive, generative language model that can optionally be conditioned on the visual modality. 
We restrict ourselves to such models, which approximate $\Pi_{k=1}^N p(t_{k+1}|t_{1:k}, v)$, where $v$ is an image and $t_{1:k}$ is a sequence of language tokens.
In a zero-shot VQA setting, it is expected that $f(\cdot)$ understands that it has been given a question $q$ and should produce the correct answer $a$ to the question $q$ in the context of the image $v$ by modeling it as $p(a|v,q)$.
This setting is common when evaluating very large frozen models, such as in \cite{Yang2021AnES,Guo2022FromIT}, with the exception that in our case, $f(\cdot)$ is a vision-language model rather than a language-only model.

\subsection{Question Decomposition}
\label{sec:decomposition-defn}
Question decomposition is the task of decomposing a complex main question into one or more simpler subquestions that are logically related to the main question, answering those simpler subquestion(s), and then using the answered subquestion(s) to help in composing a final answer to the complex main question. 
This is a strategy often used by humans for problem solving.
For example, consider a human being confronted by a wild animal they have never seen before. 
To answer the main question ``\textit{does this animal pose a threat to me?}'' a human might decompose it into subquestions such as ``\textit{does the animal have sharp canine teeth?}'' and ``\textit{does the animal have forward facing eyes typical of a predator?}'' Knowing the answer to even one of these subquestions makes answering the main question much easier.

Adopting the terminology of \cref{sec:problem-setting}, the task of question decomposition consists of \textit{decomposing} a main visual question $v,q$ into one or more subquestions $(s_1, s_2, \ldots)$, answering those subquestions to obtain the decomposition $((q^\prime_1, a^\prime_1), (q^\prime_2,a^\prime_2) \ldots)$,  and then using $v,q$ together with the decomposition $((q^\prime_1, a^\prime_1), (q^\prime_2,a^\prime_2) \ldots)$ to obtain the final answer $a$.

\subsection{What makes a good subquestion?}
\label{sec:consequentialist-defn}
In \cref{sec:decomposition-defn}, we gave a definition of decompositions that is dependent on notions of "simpler" and "logically related". 
It is challenging to make these notions precise, and difficult to operationalize them to measure whether a sequence of text really is a valid subquestion according to these notions. 
To sidestep these difficulties, we adopt a consequentialist view of whether a subquestion is ``good'', following a common consequentalist tradition in artificial intelligence as a whole \cite{aima_4e}.
We evaluate the ``goodness'' of a subquestion by measuring the effect of the subquestion.
Concretely, let $v,q,a$ be a visual question triplet where $v$ is the image, $q$ is the question, and $a$ is the answer.
Let $p_f(a|v,q)$ be the probability of the ground-truth answer $a$ as assessed by a visual question answering system $f(\cdot)$.
We regard a decomposition of $v,q$ consisting of series of subquestions and their answers $((q^\prime_1, a^\prime_1), (q^\prime_2,a^\prime_2) \ldots)$  as ``good'' if $p_f(a|v,q) < p_f(a|v,q, ((q^\prime_1, a^\prime_1), (q^\prime_2,a^\prime_2) \ldots) )$, that is, if seeing the decomposition increases the probability of the ground-truth answer $a$.
In practice, we adopt a simpler criterion that takes the consequentalist definition to the limit.
\textit{We regard a decomposition as ``good'' if seeing the decomposition induces the model to produce the true ground-truth answer $a$.}
\subsection{Scope \& Limitations}
We only consider in-context learning techniques for zero-shot VQA, and do not explore full model training in this work. 
The class of model we are interested in are instruction-following vision-language models based on large language models \cite{blip2,Liu2023VisualIT,Zhu2023MiniGPT4EV}.
This excludes previous-generation vision-language models that are not based on multi-billion parameter instruction-tuned language models \cite{Li2022BLIPBL,Khan2022SingleStreamMA,Li2021AlignBF,Lu2022UnifiedIOAU,wang2022ofa}.
Not all datasets are suitable for exploring question decomposition, as some primarily test low-level perception skills rather than high-level reasoning skills that would benefit from a decomposition.
We thus limit our evaluation to datasets that explicitly test for high-level reasoning / knowledge-based ability.
We are few-shot for the task of \textit{visual question decomposition} but zero-shot for the task of \textit{visual question answering}.

\section{How well can models use decompositions?}
\label{sec:oraclular-exprs}

\begin{table}[]
\adjustbox{max width=\linewidth}{
\begin{tabular}{@{}lrrrrrrrr@{}}
\toprule
              & \multicolumn{4}{c}{Image + Text (3B)}                                                                                                                     & \multicolumn{4}{c}{Image+Text (13B)}                                                                                                                      \\
Decomposition & \multicolumn{1}{l}{Overall}          & \multicolumn{1}{l}{Boolean}          & \multicolumn{1}{l}{Number}           & \multicolumn{1}{l}{Other}            & \multicolumn{1}{l}{Overall}          & \multicolumn{1}{l}{Boolean}          & \multicolumn{1}{l}{Number}           & \multicolumn{1}{l}{Other}            \\ \midrule
\rowcolor[HTML]{D9D9D9} 
None (Baseline)         & 79                                   & 82.5                                 & 6.8                                  & 67.4                                 & 79.1                                 & 81.9                                 & 13.7                                 & 70.1                                 \\
\rowcolor{CornflowerBlue!50}\textbf{Oracle/Oracle }          & 88.6                                 & 91.4                                 & 40.2                                 & 79.4                                 & 89.8                                 & 92.6                                 & 45.3                                 & 80.4                                 \\
$\Delta$ w.r.t Baseline          & {\color[HTML]{34A853} \textbf{9.6}}  & {\color[HTML]{34A853} \textbf{8.8}}  & {\color[HTML]{34A853} \textbf{33.3}} & {\color[HTML]{34A853} \textbf{12.1}} & {\color[HTML]{34A853} \textbf{10.8}} & {\color[HTML]{34A853} \textbf{10.7}} & {\color[HTML]{34A853} \textbf{31.6}} & {\color[HTML]{34A853} \textbf{10.3}} \\
\textbf{Oracle/Self-Answer }       & 84                                   & 87.3                                 & 21.4                                 & 72.8                                 & 83.9                                 & 87.1                                 & 26.5                                 & 73.2                                 \\
$\Delta$ w.r.t Baseline          & {\color[HTML]{34A853} 5}             & {\color[HTML]{34A853} 4.8}           & {\color[HTML]{34A853} 14.5}          & {\color[HTML]{34A853} 5.4}           & {\color[HTML]{34A853} 4.8}           & {\color[HTML]{34A853} 5.2}           & {\color[HTML]{34A853} 12.8}          & {\color[HTML]{34A853} 3.1}           \\
\textbf{Oracle/No Answer}        & 83.3                                 & 85.9                                 & 27.4                                 & 74.9                                 & 84.1                                 & 86.9                                 & 27.4                                 & 75.2                                 \\
$\Delta$ w.r.t Baseline          & {\color[HTML]{34A853} 4.4}           & {\color[HTML]{34A853} 3.4}           & {\color[HTML]{34A853} 20.5}          & {\color[HTML]{34A853} 7.6}           & {\color[HTML]{34A853} 5.1}           & {\color[HTML]{34A853} 5}             & {\color[HTML]{34A853} 13.7}          & {\color[HTML]{34A853} 5.1}           \\
\textbf{Oracle/Oracle (Scrambled)}  & 84.9                                 & 87.9                                 & 37.6                                 & 74.8                                 & 86                                   & 88.9                                 & 39.3                                 & 76.2                                 \\
$\Delta$ w.r.t Baseline          & {\color[HTML]{34A853} 5.9}           & {\color[HTML]{34A853} 5.4}           & {\color[HTML]{34A853} 30.8}          & {\color[HTML]{34A853} 7.4}           & {\color[HTML]{34A853} 6.9}           & {\color[HTML]{34A853} 7}             & {\color[HTML]{34A853} 25.6}          & {\color[HTML]{34A853} 6}             \\ \midrule
              & \multicolumn{4}{c}{Text (3B)}                                                                                                                             & \multicolumn{4}{c}{Text (13B)}                                                                                                                            \\
Decomposition & \multicolumn{1}{l}{Overall}          & \multicolumn{1}{l}{Boolean}          & \multicolumn{1}{l}{Number}           & \multicolumn{1}{l}{Other}            & \multicolumn{1}{l}{Overall}          & \multicolumn{1}{l}{Boolean}          & \multicolumn{1}{l}{Number}           & \multicolumn{1}{l}{Other}            \\ \midrule
\rowcolor[HTML]{D9D9D9} 
None (Baseline)         & 57.4                                 & 64.4                                 & 6                                    & 32.2                                 & 63.8                                 & 71.9                                 & 6.8                                  & 34.3                                 \\
\rowcolor{CornflowerBlue!50}\textbf{Oracle/Oracle}           & 72                                   & 75.8                                 & 37.6                                 & 58.4                                 & 81.5                                 & 85.1                                 & 45.3                                 & 69                                   \\
$\Delta$ w.r.t Baseline          & {\color[HTML]{34A853} \textbf{14.5}} & {\color[HTML]{34A853} \textbf{11.4}} & {\color[HTML]{34A853} \textbf{31.6}} & {\color[HTML]{34A853} \textbf{26.2}} & {\color[HTML]{34A853} \textbf{17.8}} & {\color[HTML]{34A853} \textbf{13.2}} & {\color[HTML]{34A853} \textbf{38.5}} & {\color[HTML]{34A853} \textbf{34.7}} \\
\textbf{Oracle/Self-Answer }       & 62.1                                 & 65.8                                 & 23.1                                 & 48.8                                 & 68                                   & 72.1                                 & 20.5                                 & 53.7                                 \\
$\Delta$ w.r.t Baseline        & {\color[HTML]{34A853} 4.6}           & {\color[HTML]{34A853} 1.4}           & {\color[HTML]{34A853} 17.1}          & {\color[HTML]{34A853} 16.7}          & {\color[HTML]{34A853} 4.3}           & {\color[HTML]{34A853} 0.2}           & {\color[HTML]{34A853} 13.7}          & {\color[HTML]{34A853} 19.4}          \\
\textbf{Oracle/No Answer}        & 64.8                                 & 68.7                                 & 21.4                                 & 50.9                                 & 75.2                                 & 79                                   & 26.5                                 & 62.2                                 \\
$\Delta$ w.r.t Baseline         & {\color[HTML]{34A853} 7.3}           & {\color[HTML]{34A853} 4.3}           & {\color[HTML]{34A853} 15.4}          & {\color[HTML]{34A853} 18.8}          & {\color[HTML]{34A853} 11.4}          & {\color[HTML]{34A853} 7.1}           & {\color[HTML]{34A853} 19.7}          & {\color[HTML]{34A853} 27.9}          \\
\textbf{Oracle/Oracle (Scrambled)} & 60.5                                 & 62.6                                 & 28.2                                 & 53.3                                 & 78.9                                 & 83.1                                 & 40.2                                 & 63.5                                 \\
$\Delta$ w.r.t Baseline         & {\color[HTML]{34A853} 3.1}           & {\color[HTML]{EA4335} -1.8}          & {\color[HTML]{34A853} 22.2}          & {\color[HTML]{34A853} 21.1}          & {\color[HTML]{34A853} 15.1}          & {\color[HTML]{34A853} 11.3}          & {\color[HTML]{34A853} 33.3}          & {\color[HTML]{34A853} 29.2}  \\   \bottomrule    
\end{tabular}
}
\caption{\small{Models are capable of using decompositions written by humans to provide more accurate answers. 
The gray \colorbox[HTML]{D9D9D9}{rows} are the baseline performance with no decomposition, and each $\Delta$ is calculated w.r.t to this baseline.
\textcolor{CornflowerBlue}{Oracle/Oracle} rows denoting oracle subquestions/oracle answers, have the highest $\Delta$.
``Self-Answer'' means the model answered oracular subquestions itself, and ``No Answer'' indicates the answer was left out entirely. Image+Text indicates a vision-language model (BLIP-2) was tested with multimodal inputs, while Text indicates the corresponding language model inside BLIP-2 (FLAN-T5) was tested with text only inputs. Validation split of VQA Introspect is the dataset (22k reasoning questions with their associated decompositions).}}
\label{tab:oracular}
\end{table}

Our goal in this section is to understand the ability of vision-language models based on large language models to \textit{consume} decompositions.
The hypothesis we test is:\textbf{ \textit{When provided with gold-standard decompositions on a VQA task, a model's error rate should be lower than without the gold-standard decompositions.}}
Evaluating this hypothesis presents a number of challenges.
First, how can we obtain a set of decompositions that are apriori ``known to be good''?
Second, how should the model be fed the decompositions?

To find a source of apriori ``good'' decompositions, we turn to the literature on internal consistency in visual question answering.
To probe consistency in question answering systems, several datasets \cite{Gokhale2020VQALOLVQ,vqa_introspect,Shah2019CycleConsistencyFR} have been proposed.
A particularly relevant case of such a dataset is VQA-Introspect \cite{vqa_introspect}, which probes consistency along a reasoning-perception axis.
\citet{vqa_introspect} annotate each question in the VQAv2\cite{Goyal2016MakingTV} validation set as a high-level ``reasoning'' question or a low-level ``perception'' question.
For each ``reasoning`` question, \citet{vqa_introspect} write 1-3 ``perceptual'' subquestions which are implied by the reasoning question.
For example, given a high-level reasoning question such as ``Can I eat this banana?'' a model that says ``yes'' should also reply ``yellow'' to the low-level perception question ``what is the color of the banana?''
We propose to use the low-level perception questions and answers written for the high-level reasoning questions as an \textit{oracular} decomposition for the high-level reasoning question, on the basis that the low-level perception questions are simpler than the high-level reasoning question, entail the answer for the high-level reasoning question, and are written by humans.

The second challenge lies in using a decomposition consisting of a series of subquestions and answers $((q^\prime_1, a^\prime_1), (q^\prime_2, a^\prime_2) \ldots)$  alongside a main visual question $(v,q)$.
Recall that we cannot train the model $f(\cdot)$ being used for the visual question answering task, and for any arbitrary model, it is unknown whether the model has ever seen the \textit{exact} task of decomposition-aided visual question answering.
Thus, we rely on the in-context learning ability \cite{Min2022RethinkingTR,Xie2021AnEO} of large language models to learn to perform the tasks we require from a demonstration of the task.
We handcraft a simple prompt to contain a main visual question $v,q$ from the VQAv2 validation set, along with one human-written oracular subquestion and human-written answer $q^\prime$, $a^\prime$ for the main question $v,q$ extracted from VQA-Introspect. 
The prompt is simply 
\begin{lstlisting}[language=Python,
                   basicstyle=\footnotesize\fontfamily{zi4}\selectfont,
                   keywordstyle=\color{blue},
                   commentstyle=\color{olive},
                   stringstyle=\color{BrickRed},
                   numberstyle=\tiny,
                   numbersep=5pt,
                   mathescape=true,
                   breaklines=true,
                   escapeinside=||,
                   backgroundcolor=\color{white},
                   showstringspaces=false,
                   rulecolor=\color{darkgray}]
exemplar = "Context: is the sky blue? no. are there clouds in the sky? yes. Question: what weather is likely? Short answer: rain"
prompt = exemplar + "Context: {subquestion}? {subanswer}. Question: {question}? Short answer:"
\end{lstlisting}

\textbf{Experiments \& Discussion} 
We use BLIP-2 \cite{blip2}models based on the instruction-tuned FLAN-T5\cite{Chung2022ScalingIL} in 3B and 13B sizes. 
Experiments are run on a combination of A6000s and TPUv3s, on the VQA-Introspect validation set containing 22K reasoning questions and their associated decompositions.
The results are shown in \cref{tab:oracular}.
Compared to the baseline with no oracular decompositions, both the 3B/13B vision-language models and their corresponding language models show a clear ability to benefit from decompositions across a variety of question types, with numerical questions benefiting the most.
Next, we seek to gain insight into the mechanism by which decompositions aid inference.
\textbf{\textit{Is the model merely exploiting changes in surface level statistics?}}
If so, we would expect that perturbations that leave the statistics largely unchanged but significantly alter the meaning and logical structure of the oracle decomposition should not result in significantly different performance from the unaltered oracle decompositions.
We remove the answers from the decomposition so that it only contains the subquestions, and test the effect of only using the subquestions.
Compared to the oracle, there is a significant $50$\% relative decrease in improvement w.r.t to the baseline.
Most of the subquestion answers are boolean, so removing them should not significantly change content words in the prompt, though it changes the meaning of the context significantly.
Next, we allow the models to answer the subquestions themselves (Oracle/Self-Answer) rather than using the ground-truth questions.
The accuracy of all models again decreases relative to the oracle answers, suggesting the answer and question together contribute to the result.
Finally, we take the oracle subquestion+answer and scramble the words before providing them to the models.
If the model is merely exploiting surface level statistics, the performance difference between the scrambled oracular decompositions and the original decompositions should be minimal, as the words are all the same.
Again, we observe a significant drop compared to the original decompositions, suggesting that the models \textit{are not merely exploiting changes in the surface level statistics}.
Furthermore, human-written decompositions help in almost all cases over the no-decomposition baseline.
\textbf{Note:} \textit{See supplement for complete experimental details for all experiments}.

\section{Can models produce effective decompositions?}
\label{sec:decomp-production}
\begin{figure}
    \centering
    \includegraphics[width=\linewidth]{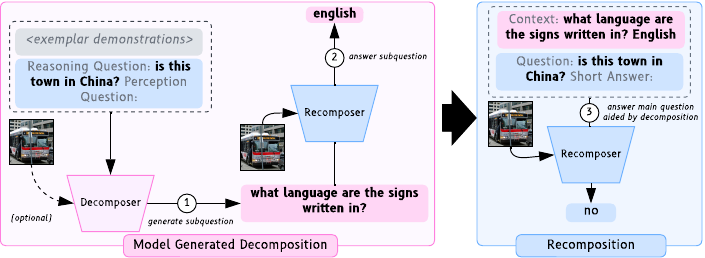}
    \caption{\small{The procedure we use to generate a decomposition and use it as additional guidance during zero-shot VQA. The recomposer can be any question answering model, and the decomposer can be any generative language model, and some models can perform both roles, leading to self-talk. In experiments, we test if various decomposer candidates can learn to write effective subquestions purely from seeing a demonstration of the task.}}
    \label{fig:producing_decompositions}
\end{figure}

\begin{table}[]
\adjustbox{max width=\linewidth}{
\begin{tabular}{@{}lcrrrrrrrrrrrrl@{}}
\toprule
                                 &                 & \multicolumn{3}{c}{A-OKVQA\cite{aokvqa}}                                                     & \multicolumn{3}{c}{ArtVQA\cite{artvqa}}                                                     & \multicolumn{3}{c}{OK-VQA\cite{okvqa}}                                                      & \multicolumn{3}{c}{SLAKE\cite{slake}}                                                      &            \\ \cmidrule(lr){3-14}
VQA Model                       & Decomposer      & \multicolumn{1}{l}{\textcolor{ForestGreen}{E$_{CR}\uparrow$}} & \multicolumn{1}{l}{\textcolor{Red}{E$_{IC}\downarrow$}} & \multicolumn{1}{l}{Err} & \multicolumn{1}{l}{\textcolor{ForestGreen}{E$_{CR}\uparrow$}} & \multicolumn{1}{l}{\textcolor{Red}{E$_{IC}\downarrow$}} & \multicolumn{1}{l}{Err} & \multicolumn{1}{l}{\textcolor{ForestGreen}{E$_{CR}\uparrow$}} & \multicolumn{1}{l}{\textcolor{Red}{E$_{IC}\downarrow$}} & \multicolumn{1}{l}{Err} & \multicolumn{1}{l}{\textcolor{ForestGreen}{E$_{CR}\uparrow$}} & \multicolumn{1}{l}{\textcolor{Red}{E$_{IC}\downarrow$}} & \multicolumn{1}{l}{Err} & Params \\ \midrule
\multirow{6}{*}{BLIP2 (3B)}  & Text   & \textbf{12.5}             & 28.12                    & 50.31                   & 7.1                       & 42.06                    & 83.15                   & \textbf{9.76}             & 31.38                    & 63.56                   & 14.12                     & 35.41                    & 66.73                   & 80.0M      \\
                                 & Text    & 10.42                     & 53.08                    & -                       & 9.56                      & 59.81                    & -                       & 9.45                      & 52.47                    & -                       & 12.15                     & 49.29                    & -                       & 250.0M     \\
                                 & Text  & 9.2                       & 30.76                    & -                       & \textbf{12.22}            & 41.12                    & -                       & 8.64                      & 29.58                    & -                       & 15.25                     & 36.83                    & -                       & 780.0M     \\ 
                                 & Text      & 7.99                      & 15.11                    & -                       & 6.06                      & 21.03                    & -                       & 7.95                      & 15.01                    & -                       & 16.38                     & 37.68                    & -                       & 3.0B       \\ \rowcolor{pink}
                                   & Image+Text  & 7.81                      & \textbf{10.9}            & -                       & 4.36                      & \textbf{13.08}           & -                       & 7.42                      & \textbf{12.29}           & -                       & 15.96                     & \textbf{28.9}            & -                       & 3.0B       \\
                                 & Text     & 9.9                       & 24.43                    & -                       & 8.05                      & 30.37                    & -                       & 9.73                      & 22.46                    & -                       & \textbf{17.09}            & 39.94                    & -                       & 11.0B      \\ \midrule
\multirow{6}{*}{BLIP2 (11B)} & Text   & 11.52                     & 33.44                    & 46.99                   & 9                         & 42.52                    & 83.15                   & 11.3                      & 34.85                    & 60.31                   & 19.12                     & 30.42                    & 77.38                   & 80.0M      \\
                                 & Text    & 8.92                      & 60.63                    & -                       & 9.94                      & 49.07                    & -                       & 9.6                       & 58.16                    & -                       & 18.15                     & 48.75                    & -                       & 250.0M     \\
                                 & Text   & 10.22                     & 36.57                    & -                       & \textbf{12.12}            & 40.65                    & -                       & 11.07                     & 35.5                     & -                       & 15.35                     & 30.83                    & -                       & 780.0M     \\
                                 & Text      & 10.78                     & \textbf{20.59}           & -                       & 8.33                      & \textbf{19.63}           & -                       & 9.73                      & \textbf{15.43}           & -                       & 19.85                     & 35.42                    & -                       & 3.0B       \\ \rowcolor{pink}
                                 & Image+Text & \textbf{14.13}            & 26.36                    & -                       & 10.61                     & 21.03                    & -                       & 13.54                     & 25.06                    & -                       & \textbf{20.71}            & \textbf{30.42}           & -                       & 11.0B      \\
                                 & Text     & 12.45                     & 30.64                    & -                       & 8.05                      & 28.97                    & -                       & \textbf{12.42}            & 27.11                    & -                       & 18.51                     & 32.5                     & -                       & 11.0B      \\ \bottomrule
\end{tabular}
}
\caption{\small{Models of drastically different sizes and multimodal capability can produce effective subquestions, as measured by \textcolor{ForestGreen}{E$_{CR}$} in \cref{eq:error_correction_rate}, their ability to correct errors in a VQA Model.
However, subquestions produced by larger models are less likely mislead the consuming VQA model, as measured by \textcolor{Red}{E$_{IC}$} in \cref{eq:error_induction_rate}.
``Text'' indicates a language-only decomposer, while ``Image+Text'' indicates a vision-language decomposer.
``Params'' refers to the parameters of the decomposer.
A pink highlight indicates when the decomposer and vqa model are the same (the model is talking to itself).
}}
\label{tab:error-correction}
\end{table}
In this section, we conduct experiments to answer the following research questions: 
\begin{enumerate}
    \item Can language models $\leq 13$B parameters learn to produce effective decompositions purely through demonstrations?
    \item Is question decomposition mostly a linguistic ability, or is being able to see the image important?
\end{enumerate}
Recall that a decomposition of a visual question $v,q$ is a series of one or more \textit{subquestions} $((q^\prime_1, a^\prime_1), (q^\prime_2, a^\prime_2) \ldots)$ 
and their answers, with the constraint that the subquestions and answers should have the property that $p_f(a|v,q) < p_f(a|v,q, ((q^\prime_1, a^\prime_1), (q^\prime_2, a^\prime_2) \ldots) )$ where $p_f(\cdot)$ represents probability assessed by a given vision-language model $f(\cdot)$ of the ground-truth answer $a$.
We simplify this task to the task of producing a \textit{single} subquestion $q^\prime$ given a main visual question $v,q$, and denote the process of decomposition with an arbitrary autoregressive language model $g(\cdot)$ as $d_g(v,q) \rightarrow q^\prime$.
We hereafter refer to the model $g(\cdot)$ that generates the decomposition as the \textit{decomposer}.
The subquestion is then answered by the vision-language model $f(v,q^\prime)=a^\prime$ to produce the subquestion-answer pair $(q^\prime, a^\prime)$.
We call the question answering model the \textit{\textit{recomposer}.}
We then measure the effectiveness of the decomposition by measuring the \textit{error correction rate}:
\begin{equation}
    \label{eq:error_correction_rate}
    \mathrm{E}_{CR} = \frac{\sum_{i=1}^N \vmathbb{1}[f(v_i,q_i) \neq a_i \wedge f(v_i, q_i, (q^\prime_i, a^\prime_i)) = a_i]}{\sum_{i=1}^N \vmathbb{1}[f(v_i,q_i) \neq a_i]}
\end{equation}
where $(v_i, q_i, a_i)$ represent the $i$-th image, question, and ground-truth answer respectively, and $q^\prime_i, a^\prime_i$ represent a subquestion generated by the decomposer model and the answer predicted for the subquestion by the recomposer (VQA) model, and $\vmathbb{1}[cond]$ is an indicator function that is equal to 1 when $cond$ is true and 0 otherwise.
Simply put, E$_{CR}$ measures the number of instances on which $f(\cdot)$ initially predicted a wrong answer, but switched to the correct answer after seeing the decomposition generated by $g(\cdot)$.
Alternatively, this can be understood as the effectiveness of a decomposer model at correcting the errors of the recomposer model.
The error induction rate E$_{IC}$ is the opposite:
\begin{equation}
    \label{eq:error_induction_rate}
    \mathrm{E}_{IC} = \frac{\sum_{i=1}^N \vmathbb{1}[f(v_i,q_i) = a_i \wedge f(v_i, q_i, (q^\prime_i, a^\prime_i)) \neq a_i]}{\sum_{i=1}^N \vmathbb{1}[f(v_i,q_i) = a_i]}
\end{equation}

and measures how often the produced decompositions flipped an answer that was initially correct to an incorrect answer.
The decomposer can be the same as the recomposer if the model can do both tasks by following different prompts, as in the case of instruction-tuned models \cite{Wei2021FinetunedLM}.

\textbf{Experiments \& Discussion} We use BLIP-2 \cite{blip2} based on the FLAN-T5\cite{Chung2022ScalingIL} as the question answering model (recomposer).
For the decomposers, we use FLAN-T5\cite{Chung2022ScalingIL} models ranging in size from $80$M parameters to $11$B parameters, as well as the BLIP-2 models themselves.
We use four VQA datasets from three domains: ArtVQA\cite{artvqa} (art), SLAKE\cite{slake} (medical), and A-OKVQA\cite{aokvqa} and OKVQA \cite{okvqa} (external knowledge VQA on natural images).
We then carry out the procedure illustrated in \cref{fig:producing_decompositions} for each combination of decomposer, recomposer, and dataset.
We handcraft three demonstrations of writing a subquestion for a question, in the form ``\textit{Reasoning Question: <question>? Perception Question: <subquestion>?}''
For each $v,q$ pair in a dataset, we prompt the decomposer with the demonstration, followed by the question $q$ as in \cref{fig:producing_decompositions}, and measure E$_{CR}$ as in \cref{eq:error_correction_rate} and E$_{IC}$ as in \cref{eq:error_induction_rate} for each dataset.
We show the results in \cref{tab:error-correction}.
We find that \textbf{\textit{yes, language models $\leq13\mathrm{B}$ parameters can learn to produce effective decompositions just by viewing examples}}.
Decomposer size correlates positively with E$_{CR}$ ($R^2=0.344$), and negatively with E$_{IC}$ ($R^2=.273$) and the correlations are significant at $\alpha=0.05$ across a larger collection of eight datasets used in \cref{sec:selective_decomposition}.
A human examination of the ``subquestions'' produced by smaller models shows that many of them are gibberish and not properly formed questions at all.
Despite this, they surprisingly manage to maintain an $E_{CR}$ that is sometimes higher than larger models.
Finally, \textbf{the ability to decompose questions in the evaluated datasets \textit{may} be a primarily linguistic ability} in that it is possible to ask effective subquestions about an image without being able to see the image, and the difference in effectiveness between the Image+Text BLIP-2 models and the text-only FLAN-T5 models of a similar size is on average $\approx 10$\% of the base error rate (but this may not be true of other VQA datasets).

\section{Selective Decomposition Works Better Than Naive Decomposition}
\label{sec:selective_decomposition}
One problem shows up in \cref{sec:decomp-production}, which is that applying decompostions to \textit{every} question can hurt performance, by flipping answers that were initially correct to be incorrect.
If we were able to decompose only wrong answers, we would always see a net gain in performance due to the error correction of decompositions.
However, in a realistic setting, we do not know apriori that our answers are wrong, and thus run the risk of flipping an answer that was initially correct to be incorrect by applying a decomposition that is misleading.
We call this the \textit{second-guessing} problem.

\begin{wrapfigure}{l}{0.4\linewidth}
\vspace{-3mm}
\begin{lstlisting}[frame=single,
                   basicstyle=\footnotesize\ttfamily,
                   mathescape=true,
                   escapeinside=||]
|$v$|: Image
|$q$|: Question
|$\tau$|: Confidence Threshold

|\color{gray}{Attempt an initial answer.}|
|$\hat{a}$|, |$p(\hat{a})$| = recomposer(|$v$|, |$q$|)
|\color{gray}{Selectively decompose $q$ if}|
|\color{gray}{answer $\hat{a}$ is uncertain.}|
if |$p(\hat{a})$| <= |$\tau$|: 
    |$q^\prime$| = decomposer(|$v$|,|$q$|)
    |$a^\prime$| = recomposer(|$v$|, |$q^\prime$|)
    |\color{gray}{Answer again.}|
    |$\hat{a}$| = recomposer(|$v$|,|$q$|,|$q^\prime, a^\prime$|)
\end{lstlisting}
\vspace{-4mm}
\caption{\small{Pseudocode for selective decomposition.}}
\label{code:selective-decomposition}
\end{wrapfigure}

To deal with the second-guessing problem, we propose following an intuitive human strategy: stick with your initial answer on questions you are confident about, and only second-guess (apply a decomposition) for questions you are not confident about.
Language models can be surprisingly well calibrated\cite{Kadavath2022LanguageM}, meaning that the probability they assess to an output sequence they produce is often well-correlated with the probability that the produced output sequence is the ``correct'' one for a given task.
We make use of this property to treat visual question answering as a selective prediction \cite{Whitehead2022ReliableVQ} task, using the language models's confidence as a decision score to determine whether we should apply a decomposition to a instance or stick with the original answer.

\begin{figure}
    \centering
    \includegraphics[width=\linewidth]{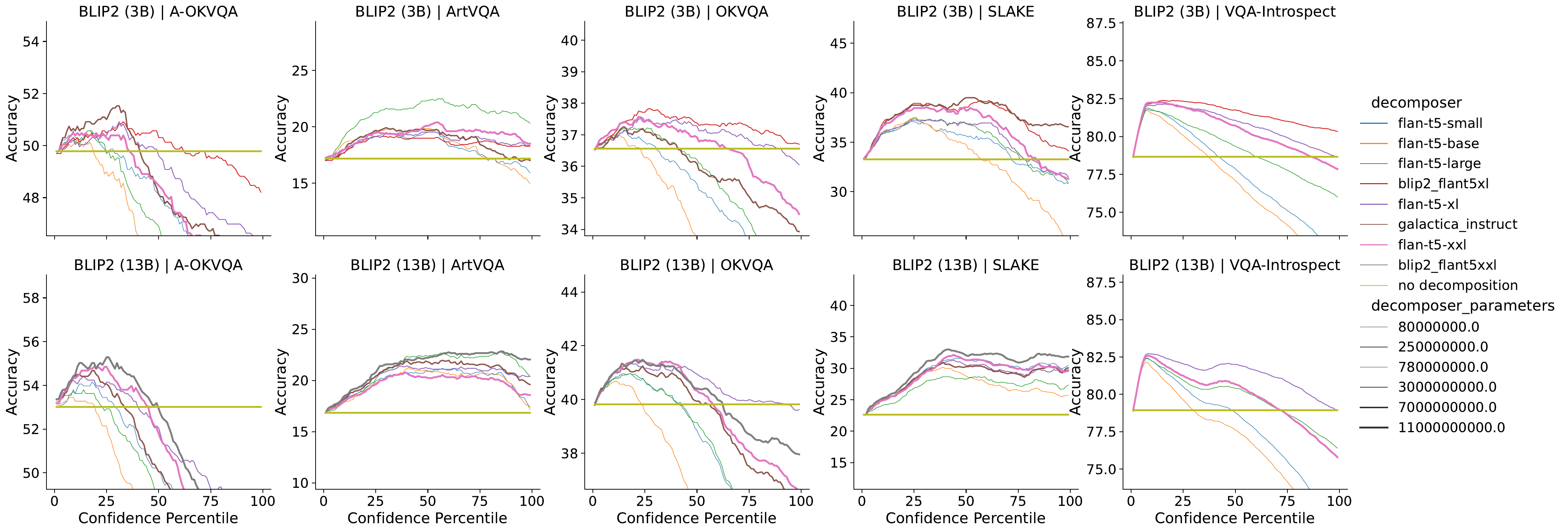}
    \caption{\small{Selective decomposition mitigates the problem of misleading decompositions. We decompose questions based on model confidence in the initial answer, and show how accuracy initial rises past the baseline as the model mostly second guesses wrong answers, and then drops below the baseline with no decompositions (horizontal line) if too many questions initially answered correctly are second-guessed.}}
    \label{fig:decomposition_curve}
\end{figure}

We describe the algorithm in pseudocode in \cref{code:selective-decomposition}.
The \textit{selective decomposition} procedure transforms VQA from a single-step task to a two-step task given a decomposer model, a recomposer model, a confidence threshold $\tau$, and a visual question pair $v,q$. 
An initial answer $\hat{a}$ and confidence $p(\hat{a})$ is solicited from the recomposer model.
If $p(\hat{a}) < \tau$, the decomposition procedure is invoked, and a subquestion and answer pair $(q^\prime, a^\prime)$ are generated by the decomposer and recomposer working together.
The recomposer model is then allowed to ``second-guess'' the inital answer $\hat{a}$ with the decomposition $(q^\prime, a^\prime)$ as additional context.
The decomposer and recomposer can be the same model or different models. 
We experiment with both scenarios.
This introduces an extra hyperparameter $\tau$ into the inference procedure.

\textbf{Experiments \& Discussion} In \cref{fig:decomposition_curve}, we show the effect of different values of $\tau$ on the accuracy of selective decomposition with several decomposers.
Across all datasets and all models, there is a wide range of $\tau$ (expressed as percentiles) for which selective decomposition improves predictive accuracy.
At the same time, we clearly demonstrate the \textit{second guessing} problem in \cref{fig:decomposition_curve}.
Decomposing \textit{every} question often eventually leads to lower accuracy than decomposing no questions at all, because hallucinations and misleading decompositions can flip an initially correct answer to an incorrect answer.
\begin{wrapfigure}{l}{0.4\linewidth}
\centering
\vspace{-4mm}
        \includegraphics[width=\linewidth]{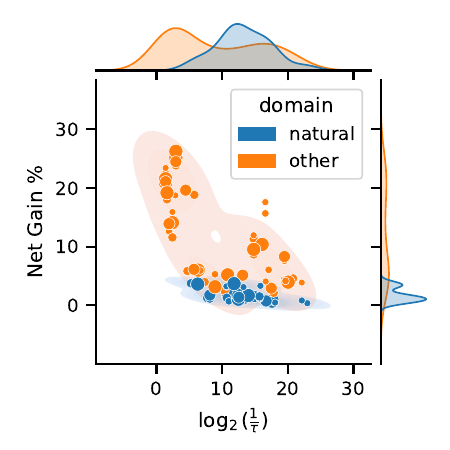}
        \vspace{-6mm}
        \caption{\small{Decomposition is more effective on non-natural image domains, and models are also less confident in these domains. Size of circles is proportional to parameter count.}}
        \vspace{-3mm}
        \label{fig:threshold_net_gain_distribution}
\end{wrapfigure}
In \cref{tab:natural_domain_net_gain,tab:other_domain_net_gain}, we show the highest possible net gain achieved by selective question decomposition on three domains by different decomposers.
Selective decomposition consistently improves predictive accuracy regardless of the decomposer and domain.
\textit{\textbf{Net gains are larger on datasets (passes t-test with $\alpha=0.05$) containing non-naturalistic images and specialized domains (e.g. medical) than they are on domains containing natural images.}}
The mean optimal surprisal $I(\tau)$ for second-guessing answers is lower for non-natural domains ($\mu_{I(\tau)}=13.2$ for the natural image datasets vs $\mu_{I(\tau)}=9.0$ for the medical and art datasets, confirmed by t-test at $\alpha=0.5$).
We further visualize this in \cref{fig:threshold_net_gain_distribution}.
\textit{This matches our expectations: you should second guess yourself on domains you understand poorly more than on domains you understand well.}
A linear regression fit shows that larger decomposers correlate with larger net gains ($R^2$=0.365, $R^2$=0.342 for natural image domains and medical / art domains respectively, t-test with $\alpha$=0.05).
 
We reformulate Winoground as a VQA task by turning each caption into a question with a boolean yes / no answer (does ``<caption>'' describe the image?) on which chance accuracy is 50$\%$.
As visible in \cref{tab:natural_domain_net_gain}, all BLIP-2 models perform below random chance, in agreement with previous results on Winoground showing that it is extremely difficult for vision-language models.
Surprisingly, after decompositions produced by the relatively FLAN-T5-small/base models (80M/200M parameters), the performance of BLIP-2 (13B) rises to significantly above chance ($+18$\%).
Upon inspection, many of the decompositions produced by the model appear to be gibberish, yet remarkably, induce the much larger 13B BLIP-2 model to correct over $30$\% of its initially wrong answers.

\begin{table}[]
\resizebox{\textwidth}{!}{%
\begin{tabular}{@{}lllrrrrrrrrrrrrrrrr@{}}
\toprule
 &
  \multicolumn{2}{c}{Decomposer} &
  \multicolumn{4}{c}{\textcolor{Purple}{AokVQA}\cite{aokvqa}} &
  \multicolumn{4}{c}{\textcolor{Purple}{OkVQA}\cite{okvqa}} &
  \multicolumn{4}{c}{\textcolor{TealBlue}{VQA Introspect}\cite{vqa_introspect}} &
  \multicolumn{4}{c}{\textcolor{TealBlue}{WinogroundVQA}\cite{winoground}} \\
  \cmidrule(l{0.5em}r{0.5em}){2-3}
\cmidrule(l{0.5em}r{0.5em}){4-7}
\cmidrule(l{0.5em}r{0.5em}){8-11}
\cmidrule(l{0.5em}r{0.5em}){12-15}
\cmidrule(l{0.5em}r{0.5em}){16-19}
VQA &
  Type &
  Params &
  \multicolumn{1}{l}{Acc} &
  \multicolumn{1}{l}{$\%\uparrow$} &
  \multicolumn{1}{l}{$\eta$} &
  \multicolumn{1}{l}{$I(\tau)$} &
  \multicolumn{1}{l}{Acc} &
  \multicolumn{1}{l}{$\%\uparrow$} &
  \multicolumn{1}{l}{$\eta$} &
  \multicolumn{1}{l}{$I(\tau)$} &
  \multicolumn{1}{l}{Acc} &
  \multicolumn{1}{l}{$\%\uparrow$} &
  \multicolumn{1}{l}{$\eta$} &
  \multicolumn{1}{l}{$I(\tau)$} &
  \multicolumn{1}{l}{Acc} &
  \multicolumn{1}{l}{$\%\uparrow$} &
  \multicolumn{1}{l}{$\eta$} &
  \multicolumn{1}{l}{$I(\tau)$} \\ \midrule
\multirow{7}{*}{3B} &
  T &
  80M &
  \multirow{7}{*}{49.69} &
  0.79 &
  16 &
  13.57 &
  \multirow{7}{*}{36.54} &
  0.59 &
  15 &
  16.69 &
  \multirow{7}{*}{78.64} &
  3.23 &
  8 &
  10.64 &
  \multirow{7}{*}{45.81} &
  5.94 &
  94 &
  0.41 \\
 &
  T &
  250M &
   &
  0.44 &
  11 &
  17.58 &
   &
  0.46 &
  13 &
  18.05 &
   &
  3.1 &
  7 &
  13.24 &
   &
  \textbf{22} &
  99 &
  0.19 \\
 &
  T &
  780M &
   &
  0.87 &
  18 &
  12.42 &
   &
  0.67 &
  26 &
  11.02 &
   &
  3.18 &
  8 &
  10.64 &
   &
  0 &
  1 &
  8.59 \\
 &
  T &
  3B &
   &
  1.14 &
  31 &
  8.13 &
   &
  0.97 &
  23 &
  12.09 &
   &
  3.5 &
  16 &
  5.87 &
   &
  0 &
  1 &
  8.59 \\
 &
  I+T &
  3B &
   &
  1.22 &
  33 &
  7.71 &
   &
  \textbf{1.29} &
  27 &
  10.7 &
   &
  \textbf{3.75} &
  20 &
  5.24 &
   &
  0 &
  1 &
  8.59 \\
 &
  T$^\star$ &
  7B &
   &
  \textbf{1.75} &
  31 &
  8.13 &
   &
  0.73 &
  15 &
  16.69 &
   & 3.37
   & 8
   & 10.64
   &
   &
  0.06 &
  2 &
  7.4 \\
 &
  T &
  11B &
   &
  0.7 &
  11 &
  17.58 &
   &
  1.03 &
  22 &
  12.47 &
   &
  3.6 &
  14 &
  6.27 &
   &
  0 &
  1 &
  8.59 \\ \midrule
\multirow{6}{*}{13B} & 
  T &
  80M &
  \multirow{6}{*}{53.36} &
  1.14 &
  18 &
  14.91 &
  \multirow{6}{*}{39.79} &
  1.15 &
  18 &
  15.68 &
  \multirow{6}{*}{78.93} &
  3.46 &
  8 &
  11.81 &
  \multirow{6}{*}{46.5} &
  6.94 &
  95 &
  0.22 \\
 &
  T &
  250M &
   &
  0.35 &
  9 &
  22.96 &
   &
  0.81 &
  11 &
  22.13 &
   &
  3.31 &
  7 &
  15.81 &
   &
  \textbf{22.12} &
  99 &
  0.09 \\
 &
  T &
  780M &
   &
  0.52 &
  14 &
  17.93 &
   &
  1.15 &
  18 &
  15.68 &
   &
  3.46 &
  8 &
  11.81 &
   &
  0.06 &
  2 &
  3.83 \\
 &
  T &
  3B &
   &
  1.05 &
  14 &
  17.93 &
   &
  1.51 &
  18 &
  15.68 &
   &
  3.73 &
  8 &
  11.81 &
   &
  0.69 &
  92 &
  0.29 \\
 &
  T$^\star$ &
  7B &
   &
  1.48 &
  18 &
  14.91 &
   &
  1.37 &
  24 &
  12.54 &
   & 3.69
   & 8
   & 11.61
   & 
   &
  4.06 &
  99 &
  0.09 \\
 &
  T &
  11B &
   &
  1.66 &
  25 &
  10.95 &
   &
  1.66 &
  21 &
  13.98 &
   &
  3.68 &
  8 &
  11.81 &
   &
  0.06 &
  14 &
  1.64 \\ 
 &
  I+T &
  11B &
   &
  \textbf{1.92} &
  25 &
  10.95 &
   &
  \textbf{1.68} &
  24 &
  12.54 &
   & \textbf{3.81}
   & 8
   & 11.81
  &
   &
  0 &
  1 &
  5.49 \\ \bottomrule
\end{tabular}
}
\caption{\small{Increases in accuracy produced by selective decomposition at the optimal second-guessing confidence threshold $\tau$, on \textcolor{Purple}{external knowledge QA} and \textcolor{TealBlue}{visual reasoning} across several decomposers. $I(\tau) = log_2(\frac{1}{\tau})$ is the surprisal of $\tau$, and $\eta$ is the percent of the questions in the dataset above $I(\tau)$, or equivalently, below $\tau$. 
T=FLAN-T5, I+T=BLIP-2 (based on FLAN-T5) and T$^\star$=Galactica.}}
\label{tab:natural_domain_net_gain}
\end{table}
\begin{table}[]
\resizebox{\textwidth}{!}{%
\begin{tabular}{@{}lllrrrrrrrrrrrrrrrr@{}} \toprule
 &
  \multicolumn{2}{c}{Decomposer} &
  \multicolumn{4}{c}{\textcolor{Purple}{ArtVQA}\cite{artvqa}} &
  \multicolumn{4}{c}{\textcolor{TealBlue}{PathVQA}\cite{pathvqa}} &
  \multicolumn{4}{c}{\textcolor{TealBlue}{SLAKE}\cite{slake}} &
  \multicolumn{4}{c}{\textcolor{TealBlue}{VQA Rad}\cite{vqa_rad}} \\
    \cmidrule(l{0.5em}r{0.5em}){2-3}
\cmidrule(l{0.5em}r{0.5em}){4-7}
\cmidrule(l{0.5em}r{0.5em}){8-11}
\cmidrule(l{0.5em}r{0.5em}){12-15}
\cmidrule(l{0.5em}r{0.5em}){16-19}
VQA &
  Type &
  Params &
  \multicolumn{1}{l}{Acc} &
  \multicolumn{1}{l}{$\%\uparrow$} &
  \multicolumn{1}{l}{$\eta$} &
  \multicolumn{1}{l}{$I(\tau)$} &
  \multicolumn{1}{l}{Acc} &
  \multicolumn{1}{l}{$\%\uparrow$} &
  \multicolumn{1}{l}{$\eta$} &
  \multicolumn{1}{l}{$I(\tau)$} &
  \multicolumn{1}{l}{Acc} &
  \multicolumn{1}{l}{$\%\uparrow$} &
  \multicolumn{1}{l}{$\eta$} &
  \multicolumn{1}{l}{$I(\tau)$} &
  \multicolumn{1}{l}{Acc} &
  \multicolumn{1}{l}{$\%\uparrow$} &
  \multicolumn{1}{l}{$\eta$} &
  \multicolumn{1}{l}{$I(\tau)$} \\ \midrule
\multirow{7}{*}{3B} &
  T &
  80M &
  \multirow{7}{*}{17.17} &
  2.36 &
  50 &
  10.37 &
  \multirow{7}{*}{12.45} &
  \textbf{15.89} &
  98 &
  2.42 &
  \multirow{7}{*}{33.27} &
  3.86 &
  24 &
  22.12 &
  \multirow{7}{*}{11.7} &
  18.73 &
  97 &
  2.88 \\
 &
  T &
  250M &
   &
  2.68 &
  50 &
  10.37 &
   &
  11.93 &
  69 &
  14.79 &
   &
  4.05 &
  26 &
  16.62 &
   &
  17.57 &
  69 &
  16.57 \\
 &
  T &
  780M &
   &
  \textbf{5.28} &
  55 &
  8.91 &
   &
  12.61 &
  99 &
  1.91 &
   &
  4.15 &
  25 &
  19.65 &
   &
  15.66 &
  69 &
  16.57 \\
 &
  T &
  3B &
   &
  2.44 &
  50 &
  10.37 &
   &
  11.56 &
  98 &
  2.42 &
   &
  3.96 &
  38 &
  7.33 &
   &
  18.82 &
  89 &
  5.76 \\
 &
  I+T &
  3B &
   &
  2.13 &
  29 &
  17.84 &
   &
  11.34 &
  69 &
  14.79 &
   &
  5.84 &
  61 &
  4.71 &
   &
  18.06 &
  98 &
  1.6 \\
 &
  T$^\star$ &
  7B &
   &
  2.91 &
  30 &
  17.47 &
   &
  13.9 &
  99 &
  1.91 &
   &
  \textbf{6.13} &
  50 &
  5.71 &
   &
  \textbf{19.62} &
  94 &
  4.44 \\
 &
  T &
  11B &
   &
  3.15 &
  55 &
  8.91 &
   &
  14.08 &
  98 &
  2.42 &
   &
  5.18 &
  31 &
  10.83 &
   &
  19.17 &
  98 &
  1.6 \\ \midrule
\multirow{7}{*}{13B} &
  T &
  80M &
  \multirow{7}{*}{16.85} &
  4.33 &
  63 &
  12.09 &
  \multirow{7}{*}{5.56} &
  23.4 &
  99 &
  1.39 &
  \multirow{7}{*}{22.62} &
  8.95 &
  45 &
  14.45 &
  \multirow{7}{*}{5.12} &
  23.89 &
  99 &
  2.95 \\
 &
  T &
  250M &
   &
  4.41 &
  41 &
  19.4 &
   &
  18.46 &
  99 &
  1.39 &
   &
  7.54 &
  39 &
  19.48 &
   &
  23.67 &
  99 &
  2.95 \\
 &
  T &
  780M &
   &
  5.91 &
  84 &
  6.67 &
   &
  \textbf{22.62} &
  99 &
  1.39 &
   &
  6.03 &
  41 &
  17.08 &
   &
  24.02 &
  99 &
  2.95 \\
 &
  T &
  3B &
   &
  4.57 &
  37 &
  20.79 &
   &
  21.33 &
  99 &
  1.39 &
   &
  8.67 &
  44 &
  14.8 &
   &
  23.84 &
  99 &
  2.95 \\
 &
  T$^\star$ &
  7B &
   &
  5.12 &
  60 &
  13.12 &
   &
  21.07 &
  99 &
  1.39 &
   &
  8.29 &
  39 &
  19.48 &
   &
  24.47 &
  99 &
  2.95 \\
 &
  T &
  11B &
   &
  3.94 &
  39 &
  20.05 &
   &
  21.56 &
  99 &
  1.39 &
   &
  9.52 &
  44 &
  14.8 &
   &
  \textbf{26.25} &
  99 &
  2.95 \\
 &
  I+T &
  11B &
   &
  \textbf{5.98} &
  86 &
  6.35 &
   &
  20.85 &
  99 &
  1.39 &
   &
  \textbf{10.37} &
  42 &
  16.08 &
   &
  25.13 &
  99 &
  2.95 \\ \bottomrule
\end{tabular}%
}
\caption{\small{Increases in accuracy produced by selective decomposition at the optimal second-guessing confidence threshold $\tau$, across two domains (\textcolor{TealBlue}{medical}/\textcolor{Purple}{art}) and several decomposers. $I(\tau) = log_2(\frac{1}{\tau})$ is the surprisal of $\tau$, and $\eta$ is the percent of the questions in the dataset above $I(\tau)$, or equivalently, below $\tau$.
T=FLAN-T5, I+T=BLIP-2 (based on FLAN-T5) and T$^\star$=Galactica.
}}
\label{tab:other_domain_net_gain}
\end{table}

\section{Literature Review}
Task decomposition \cite{Wang2022SelfConsistencyIC,Wei2022ChainOT,Zhou2022LeasttoMostPE,Khot2022DecomposedPA} improves the performance of large language models on zero-shot reasoning tasks.
The only work so far to apply similar techniques for VQA is MM-CoT \cite{zhang2023multicot}, but it does not explore task decomposition with large vision-language models, choosing to finetune a smaller model instead.
The ability to use zero-shot task decompositions may be a property of model scale, emerging at 60-200B parameters \cite{Wei2022EmergentAO}, or may be a property of large-scale pretraining on code \cite{madaan-etal-2022-language}.
Such large vision-language models have only been developed recently due to advances in vision-language alignment.
The prevailing paradigm in vision-language pretraining was to build vision-language models atop (relatively) small language models \cite{Li2022BLIPBL,Khan2022SingleStreamMA,Li2021AlignBF,Lu2022UnifiedIOAU,Khan_2023_CVPR} below 1B parameters.
Meanwhile, language models were being scaled from 3B-175B parameters \cite{Taylor2022GalacticaAL,Zhang2022OPTOP,Chung2022ScalingIL,Touvron2023LLaMAOA,Zeng2022GLM130BAO}, with each model family having at least one representative with $>10$B parameters.
Because vision-language pretraining typically requires full model training, aligning these multi-billion parameter models to the visual modality was prohibitively computationally expensive.
However, recent discoveries \cite{Khan2023ContrastiveAO,Merullo2022LinearlyMF} motivated by earlier work with frozen models \cite{frozen} have shown that the representation spaces of vision models and large-language models are surprisingly close, and rough alignment can be achieved with adapters \cite{magma} or linear mapping layers while keeping the language model frozen, and more advanced techniques have given rise to vision-LLMs \cite{Liu2023VisualIT,Zhu2023MiniGPT4EV,blip2}.
Our work is closely related to the visual question generation paradigm of \cite{Banerjee2020WeaQAWS,Changpinyo2022AllYM,Guo2022FromIT}.
However, we direct our question generation to focus on decompositions rather than general questions.

\vspace{-4mm}
\section{Conclusion}
\vspace{-3mm}
We show that question decomposition is already a viable strategy that can be used to implement a more natural approach to VQA. 
Without any training, instruction-tuned VLMs can learn to produce and use decompositions from a few demonstrations of the task.
This approach has many possible future directions.
For example, we only consider two-step approaches for visual question answering, where we ``hardcode'' the depth of the decomposition.
A natural next step would be to extend the two-step approach to a multi-step approach, which remains unexplored for large vision-language models in an open-world visual question answering setting.
Second, in-context learning has limitations.
Would models benefit from being trained to produce and consume decompositions?

\bibliography{neurips}
\bibliographystyle{unsrtnat}
\clearpage
\appendix
The supplementary materials includes a detailed description of implementation details for experiments (\cref{sec:experimental_details}), a description and statistics of all datasets used (\cref{sec:datasets}), an analysis of the effect of decomposer parameter count on error induction and error correction (\cref{sec:ecr_eic_analysis}), and a brief discussion of failure cases (\cref{sec:failure-cases}).
\section{Experimental Details}
\label{sec:experimental_details}
\subsection{Models}
We use BLIP-2 models built on the FLAN-T5 language model family.
We use the official weights and code from LAVIS \cite{li2022lavis} for the BLIP-2 visual encoder and Q-former.
The FLAN-T5 models used in experiments are provided by the Transformers \cite{Wolf2019HuggingFacesTS} library. 
The Galactica \cite{Taylor2022GalacticaAL} models we used are instruction-tuned\footnote{https://huggingface.co/GeorgiaTechResearchInstitute/galactica-6.7b-evol-instruct-70k} versions of the original galactica models, instruction-tuned on the Evol-Instruct-70k\cite{xu2023wizardlm} dataset.
For all models, we use the official wordpiece tokenizers associated with the model. 

\subsection{Image Preprocessing}
We use the same image preprocessing as in BLIP-2 \cite{blip2}, which is also identical to the image processing used in \cite{Li2022BLIPBL}. 
We resize the image to $224\times224$ using bicubic interpolation, followed by normalization of pixel values using $\mu=(0.48145466, 0.4578275, 0.40821073)$ and $\sigma = (0.26862954, 0.26130258, 0.27577711)$.
\subsection{Text Preprocessing}
We perform no preprocessing of the input text other than padding the batch of input tokens to the length of the largest sequence in the batch.
We use the same padding side as the FLAN-T5 models.
\subsection{Inference}
We use a batch size of 8 for all datasets and models.
We use \texttt{bfloat16}\footnote{https://cloud.google.com/tpu/docs/bfloat16} precision for FLAN-T5 models (including the FLAN-T5 models inside BLIP-2), and use half-precision (FP16) for the vision encoder inside BLIP-2. The Q-former is kept in full precision.
This follows the implementation in \cite{li2022lavis,blip2}.
We assign one model per compute device during inference, except when the decomposer and recomposer are the same model, in which case they share the same device.

\subsection{Sampling}
\subsubsection{Decomposition}
To produce decompositions, we use multinomial beam search sampling with 5 beams and a top-p of 0.95. 
We use a temperature of 1.0, a length penalty of 1.0, and a reptition penalty of 1.0.
These parameters were not optimized, and may be suboptimal. 
\subsubsection{Question Answering}
We use the same procedure to produce answers for questions with and without decompositions.
We use deterministic beam search with 5 beams, restricting the maximum length of the answer to 10 tokens and a minimum of one token.
We apply a length penalty of -1.

\subsection{Prompts}
\subsubsection{Decomposition}
We use the following template to prompt models to produce a decomposition of a reasoning question.
The prompt has two exemplars, each consisting of a high-level reasoning question with an associated low-level perceptual subquestion.
The exemplars are separated by newlines.
\begin{lstlisting}[language=Python,
                   basicstyle=\footnotesize\fontfamily{zi4}\selectfont,
                   keywordstyle=\color{blue},
                   commentstyle=\color{olive},
                   stringstyle=\color{BrickRed},
                   numberstyle=\tiny,
                   numbersep=5pt,
                   mathescape=true,
                   breaklines=true,
                   escapeinside=||,
                   backgroundcolor=\color{white},
                   showstringspaces=false,
                   rulecolor=\color{darkgray}]
template = 'Reasoning Question: is the banana ripe enough to eat? Perception Question: is the banana yellow?\nReasoning Question: is it cold outside? Perception Question: are any people wearing jackets?\nReasoning Question: {question} Perception Question:'
\end{lstlisting}

The galactica-instruct model requires a different prompt, which we describe below.
This is because any instructions given to the model have to match the format used in the instruction tuning dataset.
\begin{lstlisting}[language=Python,
                   basicstyle=\footnotesize\fontfamily{zi4}\selectfont,
                   keywordstyle=\color{blue},
                   commentstyle=\color{olive},
                   stringstyle=\color{BrickRed},
                   numberstyle=\tiny,
                   numbersep=5pt,
                   mathescape=true,
                   breaklines=true,
                   escapeinside=||,
                   backgroundcolor=\color{white},
                   showstringspaces=false,
                   rulecolor=\color{darkgray}]
template = (
            "Below is an instruction that describes a task. "
            "Write a response that appropriately completes the request.\n\n"
            "### Instruction:\n{instruction}\n\n### Response:"
        )
main_question = 'What country is this airline headquartered in?`
prompt = template.format(
            instruction=f"Write a simpler perception question that can help to answer: {main_question}"
        )
\end{lstlisting}

\subsection{Question Answering w/ Decomposition}
For question answering without a decomposition, we use the following template:
\begin{lstlisting}[language=Python,
                   basicstyle=\footnotesize\fontfamily{zi4}\selectfont,
                   keywordstyle=\color{blue},
                   commentstyle=\color{olive},
                   stringstyle=\color{BrickRed},
                   numberstyle=\tiny,
                   numbersep=5pt,
                   mathescape=true,
                   breaklines=true,
                   escapeinside=||,
                   backgroundcolor=\color{white},
                   showstringspaces=false,
                   rulecolor=\color{darkgray}]
template = 'Question: {question} Short Answer:'
\end{lstlisting}
This template is identical to that used by \cite{blip2}.
\subsection{Recomposition (Question Answering with Decomposition)}
For question answering aided by decomposition, we use the following template (same as the template in \cref{sec:decomp-production}).
We design the template based on examples from FLAN-T5's training templates \cite{Chung2022ScalingIL}.
Specifically, we use the keyword \texttt{Context:} to identify the start of the decomposition and prepend it to the simple question answering prompt above.
Our motivation for the design of this template is that it is conceptually similar to the reading comprehension question answering tasks in FLAN-T5's training data, which demarcate the paragraph to be read using the phrase \texttt{Context:}. 
We expect this similarity to make the task easier for the model.
\begin{lstlisting}[language=Python,
                   basicstyle=\footnotesize\fontfamily{zi4}\selectfont,
                   keywordstyle=\color{blue},
                   commentstyle=\color{olive},
                   stringstyle=\color{BrickRed},
                   numberstyle=\tiny,
                   numbersep=5pt,
                   mathescape=true,
                   breaklines=true,
                   escapeinside=||,
                   backgroundcolor=\color{white},
                   showstringspaces=false,
                   rulecolor=\color{darkgray}]
exemplar = "Context: is the sky blue? no. are there clouds in the sky? yes. Question: what weather is likely? Short answer: rain"
template = exemplar + "Context: {subquestion}? {subanswer}. Question: {question}? Short answer:"
\end{lstlisting}
\section{Datasets}
\label{sec:datasets}
\begin{table}[]
    \centering
    \begin{tabular}{llrrr}
\toprule
       Dataset &                  Type &  Images &  Questions &  Avg. Question Length \\
\midrule
       A-OKVQA & external knowledge qa &    1122 &       1145 &                  8.70 \\
        OK-VQA & external knowledge qa &    5033 &       5046 &                  8.09 \\
        ArtVQA &          fine art vqa &     718 &       1270 &                  6.51 \\
       VQA-RAD &           medical vqa &     314 &       2248 &                  6.51 \\
       PathVQA &           medical vqa &     832 &       6279 &                  6.26 \\
         SLAKE &           medical vqa &      96 &       1061 &                  8.11 \\
VQA-Introspect &      visual reasoning &   17495 &      22793 &                  5.93 \\
Winoground-VQA &      visual reasoning &     800 &       1600 &                 12.99 \\
\bottomrule
\end{tabular}
    \caption{Basic statistics for all eight datasets used in the paper.}
    \label{tab:dataset_statistics}
\end{table}
In \cref{tab:dataset_statistics}, we provide statistics of all datasets used in the paper. We further describe the datasets in this sections.

\textbf{Natural Image Datasets} These include A-OKVQA\cite{aokvqa}, OK-VQA\cite{okvqa}, VQA-Introspect\cite{vqa_introspect}, and Winoground\cite{winoground}.
These datasets include natural images only. 
For A-OKVQA, OK-VQA, and VQA-Introspect, the source of these images is the COCO\cite{coco2014} dataset.
While Winoground and VQA-Introspect contain mostly \textit{visual reasoning} that do not require significant external knowledge (e.g. historical facts), OK-VQA and A-OKVQA ask questions which require ``outside'' factual knowledge to answer, such as historical facts and contemporary information (e.g. which country does a specific airline operate in?). 

\textbf{Other Domains} Besides natural images, we also use datasets consisting of fine art images \cite{artvqa} and medical images.
The datasets consisting of medical images are themselves each drawn from different subdomains of medicine.
PathVQA \cite{pathvqa} contains pathology images, VQA-RAD \cite{vqa_rad} contains radiology images, and SLAKE \cite{slake} contains general medical images.

\section{E$_{CR}$ and E$_{IC}$ for all datasets}
\begin{table}[]
    \centering
    \adjustbox{max width=\linewidth}{
    \begin{tabular}{llllllllrrrllll}
\toprule
                & dataset & \multicolumn{3}{c}{aokvqa} & \multicolumn{3}{c}{okvqa} & \multicolumn{3}{c}{vqa-introspect} & \multicolumn{3}{l}{winogroundvqa} & Parameters \\
                & {} & \textcolor{ForestGreen}{E$_{CR}\uparrow$} & \textcolor{red}{E$_{IC}\downarrow$} &    Err & \textcolor{ForestGreen}{E$_{CR}\uparrow$} & \textcolor{red}{E$_{IC}\downarrow$} &    Err & \textcolor{ForestGreen}{E$_{CR}\uparrow$} & \textcolor{red}{E$_{IC}\downarrow$} &    Err & \textcolor{ForestGreen}{E$_{CR}\uparrow$} & \textcolor{red}{E$_{IC}\downarrow$} & \multicolumn{2}{l}{Err} \\
VQA Model & decomposer &                                           &                                     &        &                                           &                                     &        &                                           &                                     &        &                                           &                                     &        &            \\
\midrule
blip2-flant5xl & oracle-decomposer &                                       N/A &                                 N/A &    N/A &                                       N/A &                                 N/A &    N/A &                                     51.51 &                                8.39 &  22.07 &                                       N/A &                                 N/A &    N/A &        N/A \\
                & flan-t5-small &                                      12.5 &                               28.12 &  50.31 &                                      9.76 &                               31.38 &  63.56 &                                     39.76 &                               19.25 &  22.07 &                                     42.53 &                               38.63 &  54.38 &      80.0M \\
                & flan-t5-base &                                     10.42 &                               53.08 &  50.31 &                                      9.45 &                               52.47 &  63.56 &                                     39.34 &                               20.95 &  22.07 &                                     68.16 &                               32.19 &  54.38 &     250.0M \\
                & flan-t5-large &                                       9.2 &                               30.76 &  50.31 &                                      8.64 &                               29.58 &  63.56 &                                     35.49 &                               12.68 &  22.07 &                                     25.17 &                                46.3 &  54.38 &     780.0M \\
                & blip2-flant5xl &                                      7.81 &                                10.9 &  50.31 &                                      7.42 &                               12.29 &  63.56 &                                     40.44 &                                8.38 &  22.07 &                                     21.95 &                               30.55 &  54.38 &       3.0B \\
                & flan-t5-xl &                                      7.99 &                               15.11 &  50.31 &                                      7.95 &                               15.01 &  63.56 &                                     39.22 &                               10.23 &  22.07 &                                     34.02 &                               42.88 &  54.38 &       3.0B \\
                & galactica-instruct &                                     14.76 &                               24.25 &  50.31 &                                     10.48 &                               25.23 &  63.56 &                                     39.46 &                               12.46 &  22.07 &                                     28.16 &                               38.22 &  54.38 &       7.0B \\
                & flan-t5-xxl &                                       9.9 &                               24.43 &  50.31 &                                      9.73 &                               22.46 &  63.56 &                                     41.93 &                               12.10 &  22.07 &                                     28.97 &                               44.79 &  54.38 &      11.0B \\ \midrule
blip2-flant5xxl & oracle-decomposer &                                       N/A &                                 N/A &    N/A &                                       N/A &                                 N/A &    N/A &                                     58.47 &                               10.45 &  21.81 &                                       N/A &                                 N/A &    N/A &        N/A \\
                & flan-t5-small &                                     11.52 &                               33.44 &  46.99 &                                      11.3 &                               34.85 &  60.31 &                                     43.68 &                               21.85 &  21.81 &                                     43.54 &                               36.03 &  53.69 &      80.0M \\
                & flan-t5-base &                                      8.92 &                               60.63 &  46.99 &                                       9.6 &                               58.16 &  60.31 &                                     43.74 &                               24.20 &  21.81 &                                     69.03 &                               31.85 &  53.69 &     250.0M \\
                & flan-t5-large &                                     10.22 &                               36.57 &  46.99 &                                     11.07 &                                35.5 &  60.31 &                                     41.77 &                               14.12 &  21.81 &                                     23.75 &                               47.64 &  53.69 &     780.0M \\
                & flan-t5-xl &                                     10.78 &                               20.59 &  46.99 &                                      9.73 &                               15.43 &  60.31 &                                     46.06 &                               11.94 &  21.81 &                                     35.74 &                                41.7 &  53.69 &       3.0B \\
                & galactica-instruct &                                     12.83 &                               33.11 &  46.99 &                                     13.14 &                               29.51 &  60.31 &                                     46.80 &                               15.55 &  21.81 &                                     33.41 &                               30.09 &  53.69 &       7.0B \\
                & blip2-flant5xxl &                                     14.13 &                               26.36 &  46.99 &                                     13.54 &                               25.06 &  60.31 &                                     47.65 &                               12.46 &  21.81 &                                     28.52 &                               36.84 &  53.69 &      11.0B \\
                & flan-t5-xxl &                                     12.45 &                               30.64 &  46.99 &                                     12.42 &                               27.11 &  60.31 &                                     46.20 &                               16.17 &  21.81 &                                     28.06 &                               42.91 &  53.69 &      11.0B \\
\bottomrule
\end{tabular}
    }
    \caption{Error correction and error induction rates for all decomposers on natural image VQA datasets.}
    \label{tab:error_correction_rates_natural}
\end{table}

\begin{table}
    \centering
    \adjustbox{max width=\linewidth}{
    \begin{tabular}{lllllllllllllll}
\toprule
                & dataset & \multicolumn{3}{c}{artvqa} & \multicolumn{3}{c}{pathvqa} & \multicolumn{3}{c}{slake} & \multicolumn{3}{c}{vqa-rad} & Parameters \\
                & {} & \textcolor{ForestGreen}{E$_{CR}\uparrow$} & \textcolor{red}{E$_{IC}\downarrow$} &    Err & \textcolor{ForestGreen}{E$_{CR}\uparrow$} & \textcolor{red}{E$_{IC}\downarrow$} &    Err & \textcolor{ForestGreen}{E$_{CR}\uparrow$} & \textcolor{red}{E$_{IC}\downarrow$} &    Err & \textcolor{ForestGreen}{E$_{CR}\uparrow$} & \textcolor{red}{E$_{IC}\downarrow$} & \multicolumn{2}{l}{Err} \\
VQA Model & decomposer &                                           &                                     &        &                                           &                                     &        &                                           &                                     &        &                                           &                                     &        &            \\
\midrule
blip2-flant5xl & oracle-decomposer &                                       N/A &                                 N/A &    N/A &                                       N/A &                                 N/A &    N/A &                                       N/A &                                 N/A &    N/A &                                       N/A &                                 N/A &    N/A &        N/A \\
                & flan-t5-small &                                       7.1 &                               42.06 &  83.15 &                                     23.54 &                               39.64 &  87.55 &                                     14.12 &                               35.41 &  66.73 &                                     25.64 &                               37.26 &   88.3 &      80.0M \\
                & flan-t5-base &                                      9.56 &                               59.81 &  83.15 &                                     18.28 &                               44.63 &  87.55 &                                     12.15 &                               49.29 &  66.73 &                                     24.53 &                               44.49 &   88.3 &     250.0M \\
                & flan-t5-large &                                     12.22 &                               41.12 &  83.15 &                                      21.1 &                               46.55 &  87.55 &                                     15.25 &                               36.83 &  66.73 &                                     22.67 &                               41.83 &   88.3 &     780.0M \\
                & blip2-flant5xl &                                      4.36 &                               13.08 &  83.15 &                                     16.92 &                               37.47 &  87.55 &                                     15.96 &                                28.9 &  66.73 &                                     23.73 &                               24.71 &   88.3 &       3.0B \\
                & flan-t5-xl &                                      6.06 &                               21.03 &  83.15 &                                     19.14 &                               42.58 &  87.55 &                                     16.38 &                               37.68 &  66.73 &                                     24.99 &                               30.04 &   88.3 &       3.0B \\
                & galactica-instruct &                                      7.95 &                               38.32 &  83.15 &                                     22.76 &                               47.06 &  87.55 &                                     18.08 &                               26.35 &  66.73 &                                      26.8 &                               38.78 &   88.3 &       7.0B \\
                & flan-t5-xxl &                                      8.05 &                               30.37 &  83.15 &                                     22.65 &                               46.04 &  87.55 &                                     17.09 &                               39.94 &  66.73 &                                     25.49 &                               33.08 &   88.3 &      11.0B \\ \midrule
blip2-flant5xxl & oracle-decomposer &                                       N/A &                                 N/A &    N/A &                                       N/A &                                 N/A &    N/A &                                       N/A &                                 N/A &    N/A &                                       N/A &                                 N/A &    N/A &        N/A \\
                & flan-t5-small &                                       9.0 &                               42.52 &  83.15 &                                     27.49 &                               41.83 &  94.44 &                                     19.12 &                               30.42 &  77.38 &                                      27.0 &                               34.78 &  94.88 &      80.0M \\
                & flan-t5-base &                                      9.94 &                               49.07 &  83.15 &                                     21.92 &                                40.4 &  94.44 &                                     18.15 &                               48.75 &  77.38 &                                     26.82 &                               26.96 &  94.88 &     250.0M \\
                & flan-t5-large &                                     12.12 &                               40.65 &  83.15 &                                     26.21 &                               35.82 &  94.44 &                                     15.35 &                               30.83 &  77.38 &                                     27.29 &                               33.91 &  94.88 &     780.0M \\
                & flan-t5-xl &                                      8.33 &                               19.63 &  83.15 &                                     24.99 &                               39.54 &  94.44 &                                     19.85 &                               35.42 &  77.38 &                                     26.63 &                               21.74 &  94.88 &       3.0B \\
                & galactica-instruct &                                     10.42 &                               36.45 &  83.15 &                                      25.3 &                               42.98 &  94.44 &                                     21.92 &                                42.5 &  77.38 &                                     28.18 &                               40.87 &  94.88 &       7.0B \\
                & blip2-flant5xxl &                                     10.61 &                               21.03 &  83.15 &                                     24.33 &                               34.96 &  94.44 &                                     20.71 &                               30.42 &  77.38 &                                     28.04 &                               24.35 &  94.88 &      11.0B \\
                & flan-t5-xxl &                                      8.05 &                               28.97 &  83.15 &                                     25.28 &                               38.97 &  94.44 &                                     18.51 &                                32.5 &  77.38 &                                     29.35 &                               29.57 &  94.88 &      11.0B \\
\bottomrule
\end{tabular}

    }
    \caption{Error correction and error induction rates for all decomposers on non-natural image domains (medical and fine art VQA).}
    \label{tab:error_correction_rates_other_domains}
\end{table}

In \cref{tab:error_correction_rates_natural,tab:error_correction_rates_other_domains}, we show E$_{CR}$ and E$_{IC}$ for all decomposers and all datasets used.
We note that the oracular decompositions appear to have a similar error induction rate E$_{IC}$ as the best model-generated decompositions (BLIP2-FLANT5XL / BLIP2-FLANT5XXL), but have a noticeably higher error correction rate E$_{CR}$ of $+10$\% relative to the best model generated decompositions.
An observation from this is that the model has a limited capacity to reason from decompositions, because even human-generated, oracular decompositions mislead it roughly 8\% of the time.
Another point of note is that the instruction-tuned Galactica \cite{Taylor2022GalacticaAL} model is not significantly better at writing decompositions than the FLAN-T5 models on medical datasets, despite being trained on much more scientific data.

\section{ E$_{IC}$ Drops Faster Than E$_{CR}$ Rises}
\label{sec:ecr_eic_analysis}
\begin{figure}
    \centering
    \includegraphics[width=0.75\linewidth]{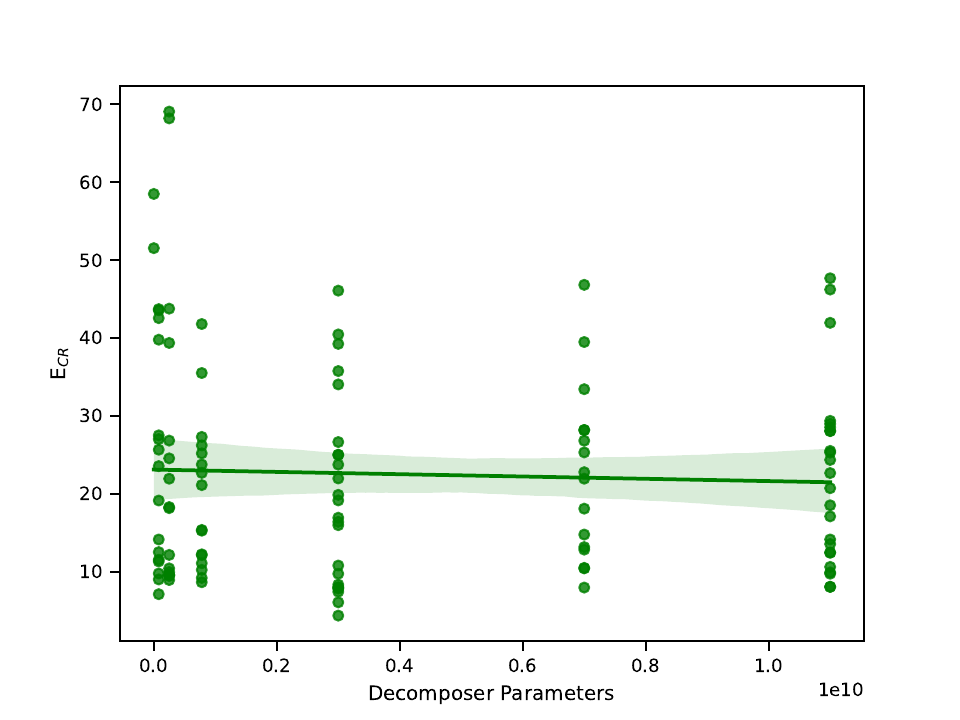}
    \caption{Error correction rate E$_{CR}$ on all datasets (except Winoground) with respect to the number of parameters in the decomposer. There is statiscally significant correlation  between the number of parameters $R^2=0.40$. The slope is $0.0215$ when the unit scale is set to $100$M parameters, corresponding to a $\approx .2\%$ increase in E$_{CR}$ for every $1$B increase in parameters.}
    \label{fig:ecr_vs_parameters}
\end{figure}

\begin{figure}
\vspace{-12mm}
    \centering
    \includegraphics[width=0.75\linewidth]{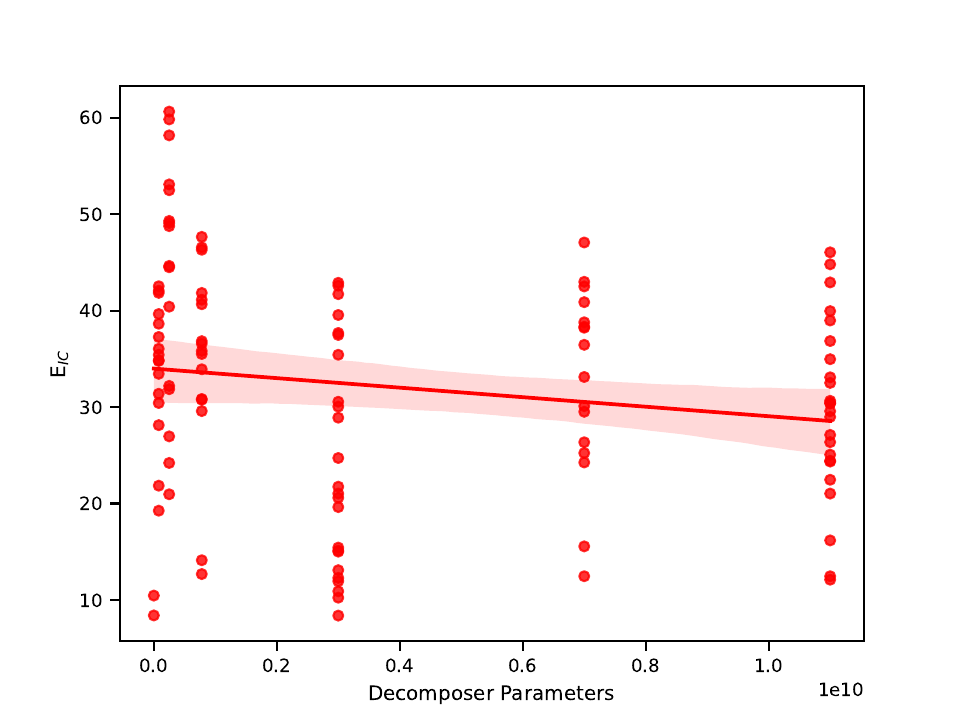}
    \caption{Error induction rate E$_{IC}$ on all datasets (except Winoground) with respect to the number of parameters in the decomposer. There is statiscally significant correlation  between the number of parameters $R^2=0.35$. The slope is $-0.07$ when the unit scale is set to $100$M parameters, corresponding to a $\approx .7\%$ increase in E$_{CR}$ for every $1$B increase in parameters.}
    \label{fig:eic_vs_parameters}
\end{figure}
In \cref{fig:ecr_vs_parameters,fig:eic_vs_parameters}, we plot the relationship between E$_{CR}$, E$_{IC}$, and parameter count of the decomposer.
We exclude Winoground from the plots because the mechanism of effect of decompositions appears to be different for Winoground.
There are statistically significant relationships (at the 95\% significance level, $\alpha=0.05$) for both E$_{CR}$ and E$_{IC}$.
E$_{IC}$ drops $.7$\% for every 1B increase in parameters, while E$_{CR}$ increases $.2$\% for every 1B increase in parameters.
This indicates that the strongest effect of scaling is to produce less misleading decompositions.
The ability to produce decompositions that correct more and more errors appears to increase more slowly with scale.

\section{Failure Cases}
\label{sec:failure-cases}
\begin{figure}
    \centering
    \includegraphics{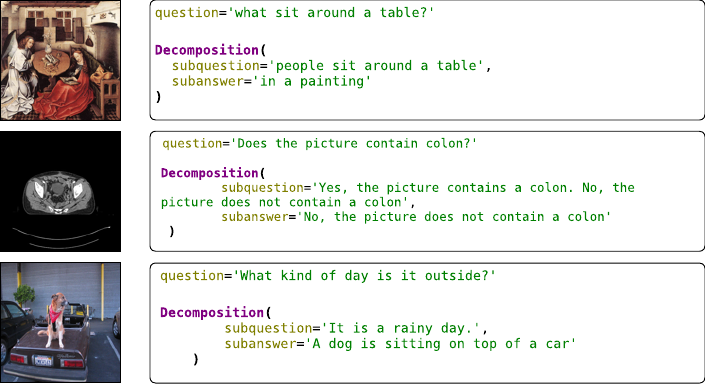}
    \caption{Examples of failure cases when attempting to produce decompositions.}
    \label{fig:failure_cases}
\end{figure}

In \cref{fig:failure_cases}, we show examples of failure cases that occur when attempting to produce decompositions.
The incidence of failure cases varies by domain and model size. 
On natural image domains and for large models (3B or more), the number of failure cases is very low.
For non-natural image domains (e.g. art), even the largest models have a high incidence of failure cases in which the produced decomposition is not even a question.
In some cases (e.g. Winoground) the failed decompositions can still correct the answer, even when they appear to be unrelated to the content of the image.
We hypothesize that there is a connection between this failure mode (apparently unrelated text results in the right answer) and the phenomenon of nonsense prompts in discrete prompt tuning \cite{rlprompt}, in which prefixing an apparently random sequence of words to a prompt results in significantly increased performance.

\clearpage

\begin{table}[]
\resizebox{\textwidth}{!}{
\begin{tabular}{@{}lllrrrrrrrrrrrrrrrr@{}}
\toprule
 &
  \multicolumn{2}{c}{Decomposer} &
  \multicolumn{4}{c}{\textcolor{Purple}{AokVQA}\cite{aokvqa}} &
  \multicolumn{4}{c}{\textcolor{Purple}{OkVQA}\cite{okvqa}} &
  \multicolumn{4}{c}{\textcolor{TealBlue}{VQA Introspect}\cite{vqa_introspect}} &
  \multicolumn{4}{c}{\textcolor{TealBlue}{WinogroundVQA}\cite{winoground}} \\
  \cmidrule(l{0.5em}r{0.5em}){2-3}
\cmidrule(l{0.5em}r{0.5em}){4-7}
\cmidrule(l{0.5em}r{0.5em}){8-11}
\cmidrule(l{0.5em}r{0.5em}){12-15}
\cmidrule(l{0.5em}r{0.5em}){16-19}
VQA &
  Type &
  Params &
  \multicolumn{1}{l}{Acc} &
  \multicolumn{1}{l}{$\%\uparrow$} &
  \multicolumn{1}{l}{$\eta$} &
  \multicolumn{1}{l}{$I(\tau)$} &
  \multicolumn{1}{l}{Acc} &
  \multicolumn{1}{l}{$\%\uparrow$} &
  \multicolumn{1}{l}{$\eta$} &
  \multicolumn{1}{l}{$I(\tau)$} &
  \multicolumn{1}{l}{Acc} &
  \multicolumn{1}{l}{$\%\uparrow$} &
  \multicolumn{1}{l}{$\eta$} &
  \multicolumn{1}{l}{$I(\tau)$} &
  \multicolumn{1}{l}{Acc} &
  \multicolumn{1}{l}{$\%\uparrow$} &
  \multicolumn{1}{l}{$\eta$} &
  \multicolumn{1}{l}{$I(\tau)$} \\ \midrule
\multirow{5}{*}{OpenFlamingo} & T         & 80M    & \multirow{5}{*}{0.96}  & 9.69         & 99     & 0         & \multirow{5}{*}{0.59}  & 9.81         & 99     & 0         & \multirow{5}{*}{5.68}  & 21           & 99     & 0.01      & \multirow{5}{*}{2.38}  & 26.69        & 99     & 0.02      \\
                              & T         & 3B     &                        & 15.28        & 99     & 0         &                        & 13.79        & 99     & 0         &                        & 35           & 99     & 0.01      &                        & 19.56        & 99     & 0.02      \\
                              & T         & 11B    &                        & 12.49        & 99     & 0         &                        & 11.41        & 99     & 0         &                        & 38.23        & 99     & 0.01      &                        & 22.62        & 99     & 0.02      \\
                              & Galactica & 7B     &                        & 9.78         & 97     & 0.01      &                        & 11.24        & 99     & 0         &                        & 35.73        & 99     & 0.01      &                        & 16.94        & 99     & 0.02      \\
                              & Falcon    & 7B     &                        & 9.52         & 99     & 0         &                        & 9.47         & 99     & 0         &                        & 31.41        & 99     & 0.01      &                        & 28.31        & 99     & 0.02      \\ \midrule
\multirow{5}{*}{InstructBLIP} & T         & 80M    & \multirow{5}{*}{24.63} & 20.96        & 99     & 0.03      & \multirow{5}{*}{36.78} & 4.6          & 99     & 0.03      & \multirow{5}{*}{75.86} & 7.68         & 47     & 0.12      & \multirow{5}{*}{12.88} & 18.69        & 99     & 0.05      \\
                              & T         & 3B     &                        & 21.22        & 99     & 0.03      &                        & 7.59         & 99     & 0.03      &                        & 6.68         & 48     & 0.12      &                        & 20.44        & 99     & 0.05      \\
                              & T         & 11B    &                        & 20.79        & 99     & 0.03      &                        & 5.87         & 99     & 0.03      &                        & 6.95         & 45     & 0.13      &                        & 16.25        & 99     & 0.05      \\
                              & Galactica & 7B     &                        & 22.1         & 99     & 0.03      &                        & 6.68         & 99     & 0.03      &                        & 7.95         & 52     & 0.11      &                        & 17.31        & 99     & 0.05      \\
                              & Falcon    & 7B     &                        & 16.51        & 99     & 0.03      &                        & 2.81         & 99     & 0.03      &                        & 4.32         & 21     & 0.23      &                        & 13.13        & 99     & 0.05      \\ 
                              \midrule
                              BLIP2 (3B) &
  Galactica &
  7B &
  \multirow{2}{*}{49.78} &
  1.75 &
  31 &
  8.13 &
  \multirow{2}{*}{36.52} &
  0.73 &
  15 &
  16.69 &
  \multirow{2}{*}{78.63} &
  3.37 &
  8 &
  10.64 &
  \multirow{2}{*}{45.69} &
  0.06 &
  2 &
  7.4 \\
 & Falcon & 7B &    & 0.7    & 18  & 12.91  &    & 0.79  & 16  & 15.87  &      & 3.35     & 8     & 10.61    &      & 1.31     & 99    & 0.19    \\ \midrule
BLIP2 (11B) &
  Galactica &
  7B &
  \multirow{2}{*}{53.19} &
  1.48 &
  18 &
  14.91 &
  \multirow{2}{*}{39.87} &
  1.37 &
  24 &
  12.54 &
  \multirow{2}{*}{78.96} &
  3.69 &
  8 &
  11.81 &
  \multirow{2}{*}{46.38} &
  4.06 &
  99 &
  0.09 \\
 & Falcon & 7B &    & 1.22   & 13  & 18.63  &    & 1.49  & 17  & 16.3   &      & 3.58     & 8     & 11.82    &      & 7.75     & 99    & 0.09   \\ \bottomrule
\end{tabular}
}\caption{Experiments with OpenFlamingo \cite{awadalla2023openflamingo}, InstructBLIP \cite{instructblip} and Falcon \cite{refinedweb} on natural image domains.}
\end{table}
\begin{table}[]
\resizebox{\textwidth}{!}{%
\begin{tabular}{@{}lllrrrrrrrrrrrrrrrr@{}} \toprule
 &
  \multicolumn{2}{c}{Decomposer} &
  \multicolumn{4}{c}{\textcolor{Purple}{ArtVQA}\cite{artvqa}} &
  \multicolumn{4}{c}{\textcolor{TealBlue}{PathVQA}\cite{pathvqa}} &
  \multicolumn{4}{c}{\textcolor{TealBlue}{SLAKE}\cite{slake}} &
  \multicolumn{4}{c}{\textcolor{TealBlue}{VQA Rad}\cite{vqa_rad}} \\
    \cmidrule(l{0.5em}r{0.5em}){2-3}
\cmidrule(l{0.5em}r{0.5em}){4-7}
\cmidrule(l{0.5em}r{0.5em}){8-11}
\cmidrule(l{0.5em}r{0.5em}){12-15}
\cmidrule(l{0.5em}r{0.5em}){16-19}
VQA &
  Type &
  Params &
  \multicolumn{1}{l}{Acc} &
  \multicolumn{1}{l}{$\%\uparrow$} &
  \multicolumn{1}{l}{$\eta$} &
  \multicolumn{1}{l}{$I(\tau)$} &
  \multicolumn{1}{l}{Acc} &
  \multicolumn{1}{l}{$\%\uparrow$} &
  \multicolumn{1}{l}{$\eta$} &
  \multicolumn{1}{l}{$I(\tau)$} &
  \multicolumn{1}{l}{Acc} &
  \multicolumn{1}{l}{$\%\uparrow$} &
  \multicolumn{1}{l}{$\eta$} &
  \multicolumn{1}{l}{$I(\tau)$} &
  \multicolumn{1}{l}{Acc} &
  \multicolumn{1}{l}{$\%\uparrow$} &
  \multicolumn{1}{l}{$\eta$} &
  \multicolumn{1}{l}{$I(\tau)$} \\ \midrule
\multirow{5}{*}{OpenFlamingo} & T         & 80M & \multirow{5}{*}{0}      & 3.39  & 99  & 0   & \multirow{5}{*}{3.44}   & 7.26  & 99  & 0.01 & \multirow{5}{*}{0.85}  & 17.34 & 99 & 0.01 & \multirow{5}{*}{6.01}   & 13.61  & 89 & 0.02 \\
                              & T         & 3B  &        & 5.83  & 99  & 0   &        & 15.81 & 99  & 0.01 &       & 27.62 & 99 & 0.01 &        & 20.11  & 98 & 0.01 \\
                              & T         & 11B &        & 5.04  & 99  & 0   &        & 16.31 & 99  & 0.01 &       & 25.82 & 99 & 0.01 &        & 20.64  & 98 & 0.01 \\
                              & Galactica & 7B  &        & 3.07  & 99  & 0   &        & 19.05 & 99  & 0.01 &       & 22.9  & 99 & 0.01 &        & 17.35  & 99 & 0.01 \\
                              & Falcon    & 5B  &        & 7.8   & 99  & 0   &        &       & N/A &      &       & 17.06 & 99 & 0.01 &        & 20.33  & 89 & 0.02 \\ \midrule
\multirow{5}{*}{InstructBLIP} & T         & 80M & \multirow{5}{*}{32.36}  & 0     & 1   & 0.9 & \multirow{5}{*}{28.06}  & 2.05  & 97  & 0.08 & \multirow{5}{*}{28.93} & 4.62  & 43 & 0.29 & \multirow{5}{*}{32.61}  & 0.89   & 52 & 0.23 \\
                              & T         & 3B  &        & 0     & 1   & 0.9 &        & 1.64  & 87  & 0.15 &       & 4.15  & 92 & 0.1  &        & 1.65   & 60 & 0.2  \\
                              & T         & 11B &        & 0     & 1   & 0.9 &        & 2.85  & 97  & 0.08 &       & 4.24  & 43 & 0.29 &        & 0.8    & 52 & 0.23 \\
                              & Galactica & 7B  &        & 0     & 1   & 0.9 &        & 2.01  & 86  & 0.15 &       & 4.52  & 56 & 0.23 &        & 1.33   & 51 & 0.23 \\
                              & Falcon    & 5B  &        & 0     & 1   & 0.9 &        & 1.24  & 76  & 0.19 &       & 3.2   & 44 & 0.28 &        & 0.85   & 50 & 0.23 \\
                              \midrule
                              \multirow{2}{*}{BLIP2 (3B)} &
  Galactica &
  7B &
  \multirow{2}{*}{17.01} &
  2.91 &
  30 &
  17.47 &
  \multirow{2}{*}{12.45} &
  13.9 &
  99 &
  1.91 &
  \multirow{2}{*}{33.36} &
  6.13 &
  50 &
  5.71 &
  \multirow{2}{*}{11.7} &
  19.62 &
  94 &
  4.44 \\
 & Falcon & 5B &    & 1.73   & 32  & 16.64  &    & 13.71   & 89   & 6.12  &    & 5.75  & 26  & 16.67  &    & 17.22   & 71  & 13.67  \\ \midrule
\multirow{2}{*}{BLIP2 (11B)} &
  Galactica &
  7B &
  \multirow{2}{*}{16.85} &
  5.12 &
  60 &
  13.12 &
  \multirow{2}{*}{5.56} &
  21.07 &
  99 &
  1.39 &
  \multirow{2}{*}{22.62} &
  8.29 &
  39 &
  19.48 &
  \multirow{2}{*}{5.12} &
  24.47 &
  99 &
  2.95 \\
 & Falcon & 5B &    & 3.39   & 41  & 19.39  &    & 20.61   & 99   & 1.39  &    & 9.52  & 88  & 2.39   &    & 23.44   & 99  & 2.91  \\ \bottomrule
\end{tabular}
}
\caption{Experiments with OpenFlamingo \cite{awadalla2023openflamingo}, InstructBLIP \cite{instructblip} and Falcon \cite{refinedweb} on non-natural image domains.}
\end{table}
\begin{figure}
    \centering
    \includegraphics[width=\linewidth]{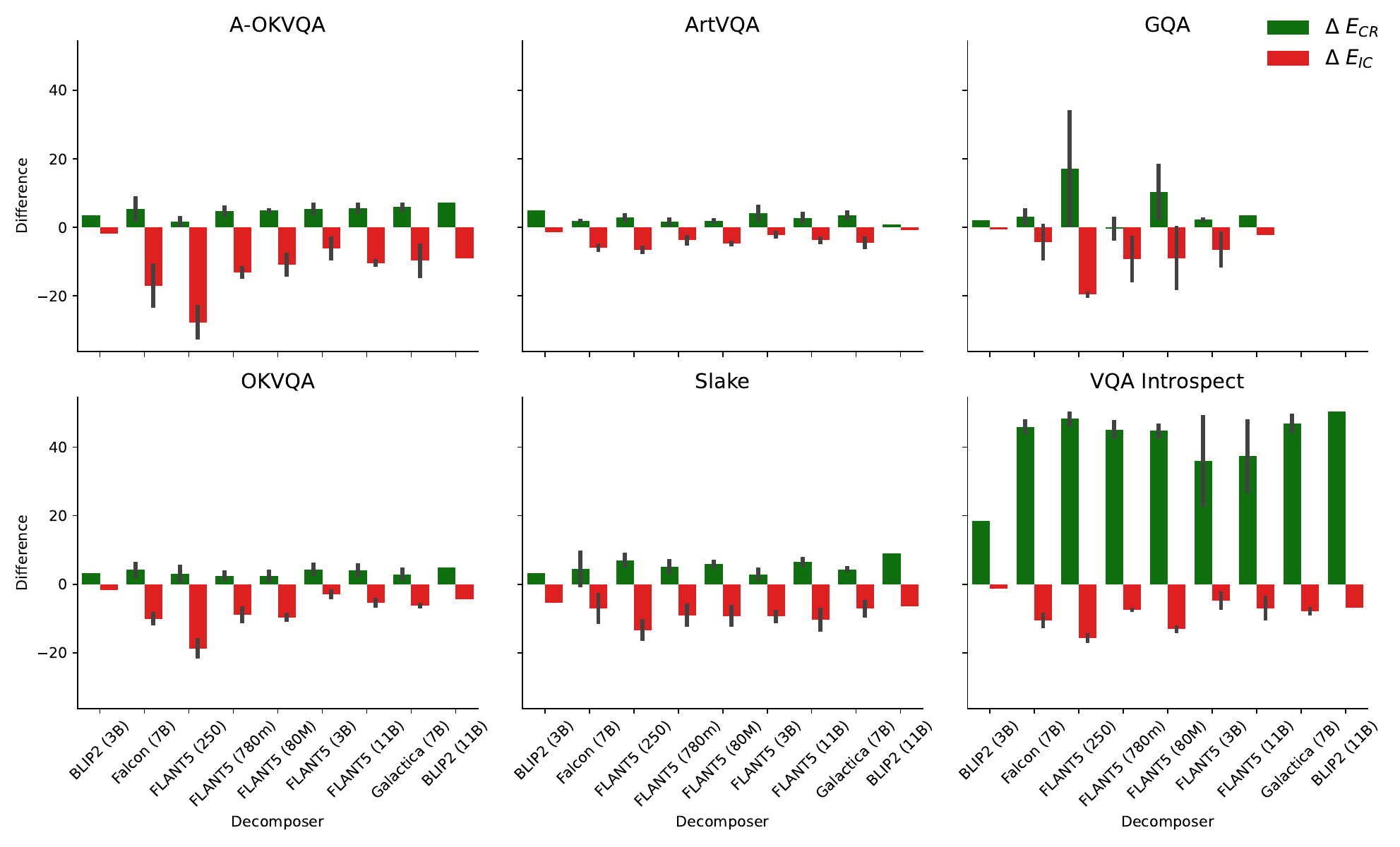}
    \caption{The change in $E_{IC}$ and $E_{CR}$ after selectively decomposing questions.}
    \label{fig:enter-label}
\end{figure}

\end{document}